\documentclass{article}

\usepackage{microtype}
\usepackage{graphicx}
\usepackage{subfigure}
\usepackage{booktabs}

\usepackage{url}
\usepackage{multibib}
\usepackage{amsmath}
\usepackage{amsthm}
\usepackage{amsfonts}
\usepackage{amssymb}
\usepackage{graphicx}
\usepackage{color}
\usepackage{xcolor}
\usepackage{graphicx}
\usepackage{multirow}
\usepackage{float}
\usepackage{placeins}
\usepackage{hyperref}

\newcommand{\minibatch}{\mathcal{M}}

\newcommand{\dkls}[3]{\mathbb{D}_{KL}^{#1}[#2 \, \|\, #3]}

\newcommand{\sg}[2]{\widehat{\nabla}_{#2} #1}

\newcommand\cut[1]{}

\newcommand{\tlambda}{\tilde{\lambda}}

\usepackage{algorithm}

\newcommand{\squishlist}{
   \begin{list}{$\bullet$}
    { \setlength{\itemsep}{0pt}      \setlength{\parsep}{3pt}
      \setlength{\topsep}{3pt}       \setlength{\partopsep}{0pt}
      \setlength{\leftmargin}{1.5em} \setlength{\labelwidth}{1em}
      \setlength{\labelsep}{0.5em} } }

\newcommand{\squishlisttwo}{
   \begin{list}{$\bullet$}
    { \setlength{\itemsep}{0pt}    \setlength{\parsep}{0pt}
      \setlength{\topsep}{0pt}     \setlength{\partopsep}{0pt}
      \setlength{\leftmargin}{2em} \setlength{\labelwidth}{1.5em}
      \setlength{\labelsep}{0.5em} } }

\newcommand{\squishend}{
    \end{list}  }
 
{}
\newtheorem{thm}{Theorem}{}
{}
{}
{}

\newcommand{\half}{\mbox{$\frac{1}{2}$}}

\newcommand{\real}{\mbox{$\mathbb{R}$}}

\newcommand{\rnd}[1]{\left(#1\right)}
\newcommand{\sqr}[1]{\left[#1\right]}

\newcommand{\myexpect}{\mathbb{E}}

\newcommand{\gauss}{\mbox{${\cal N}$}}

\newcommand{\myvec}[1]{\mbox{$\mathbf{#1}$}}
\newcommand{\myvecsym}[1]{\mbox{$\boldsymbol{#1}$}}

\newcommand{\vepsilon}{\mbox{$\myvecsym{\epsilon}$}}
\newcommand{\veta}{\mbox{$\myvecsym{\eta}$}}

\newcommand{\vmu}{\mbox{$\myvecsym{\mu}$}}

\newcommand{\vtheta}{\mbox{$\myvecsym{\theta}$}}

\newcommand{\vsigma}{\mbox{$\myvecsym{\sigma}$}}
\newcommand{\vSigma}{\mbox{$\myvecsym{\Sigma}$}}

\newcommand{\va}{\mbox{$\myvec{a}$}}
\newcommand{\vb}{\mbox{$\myvec{b}$}}

\newcommand{\vg}{\mbox{$\myvec{g}$}}
\newcommand{\vh}{\mbox{$\myvec{h}$}}

\newcommand{\vm}{\mbox{$\myvec{m}$}}

\newcommand{\vs}{\mbox{$\myvec{s}$}}

\newcommand{\vu}{\mbox{$\myvec{u}$}}

\newcommand{\vx}{\mbox{$\myvec{x}$}}

\newcommand{\vF}{\mbox{$\myvec{F}$}}
\newcommand{\vG}{\mbox{$\myvec{G}$}}
\newcommand{\vH}{\mbox{$\myvec{H}$}}
\newcommand{\vI}{\mbox{$\myvec{I}$}}

\newcommand{\vM}{\mbox{$\myvec{M}$}}

\newcommand{\vS}{\mbox{$\myvec{S}$}}

\newcommand{\Cov}{\mbox{$\mbox{Cov}$}}
\newcommand{\diag}{\mbox{$\mbox{diag}$}}

\newcommand{\calD}{\mbox{${\cal D}$}}

\newcommand{\data}{\calD}

\newcommand{\ee}{\end{equation}}
\newcommand{\bea}{\begin{eqnarray}}
\newcommand{\eea}{\end{eqnarray}}
\newcommand{\beaa}{\begin{eqnarray*}}
\newcommand{\eeaa}{\end{eqnarray*}}

\newcommand{\state}{\boldsymbol{\mathrm{s}}}
\newcommand{\action}{\boldsymbol{\mathrm{a}}}
\newcommand{\vomega}{\boldsymbol{\mathrm{\omega}}}
\newcommand{\params}{\boldsymbol{\mathrm{\theta}}}

\usepackage[accepted, nohyperref]{icml2018}

\icmltitlerunning{Bayesian Deep Learning by Weight-Perturbation in Adam}

\begin{document}

\twocolumn[
   \icmltitle{Fast and Scalable Bayesian Deep Learning by Weight-Perturbation in Adam}



\icmlsetsymbol{equal}{*}

\begin{icmlauthorlist}
\icmlauthor{Mohammad Emtiyaz Khan}{equal,rik}
\icmlauthor{Didrik Nielsen}{equal,rik}
\icmlauthor{Voot Tangkaratt}{equal,rik}
\icmlauthor{Wu Lin}{ubc}
\icmlauthor{Yarin Gal}{oxford}
\icmlauthor{Akash Srivastava}{edin}
\end{icmlauthorlist}

\icmlaffiliation{rik}{RIKEN Center for Advanced Intelligence project, Tokyo, Japan}
\icmlaffiliation{ubc}{University of British Columbia, Vancouver, Canada}
\icmlaffiliation{oxford}{University of Oxford, Oxford, UK}
\icmlaffiliation{edin}{University of Edinburgh, Edinburgh, UK}
\icmlcorrespondingauthor{Mohammad Emtiyaz Khan}{emtiyaz.khan@riken.jp}

\icmlkeywords{Bayesian deep learning, uncertainty estimation, Bayesian neural networks, variational inference, variational optimization, Adam, Vadam, Vprop, natural momentum}

\vskip 0.3in
]



\printAffiliationsAndNotice{\icmlEqualContribution} 

\begin{abstract}
   Uncertainty computation in deep learning is essential to design robust and reliable systems. Variational inference (VI) is a promising approach for such computation, but requires more effort to implement and execute compared to maximum-likelihood methods. In this paper, we propose new natural-gradient algorithms to reduce such efforts for Gaussian mean-field VI. Our algorithms can be implemented within the Adam optimizer by perturbing the network weights during gradient
evaluations, and uncertainty estimates can be cheaply obtained by using the vector that adapts the learning rate. This requires lower memory, computation, and implementation effort than existing VI methods, while obtaining uncertainty estimates of comparable quality. Our empirical results confirm this and further suggest that the weight-perturbation in our algorithm could be useful for exploration in reinforcement learning and stochastic optimization.

\end{abstract}

\section{Introduction}
\label{sec:intro}

Deep learning methods have had enormous recent success in fields where prediction accuracy is important, e.g., computer vision and speech recognition.
However, for these methods to be useful in fields such as robotics and medical diagnostics, we need to know the uncertainty of our predictions. 
For example, physicians might need such uncertainty estimates to choose a safe but effective treatment for their patients.
Lack of such estimates might result in unreliable decisions which can sometime have disastrous consequences.

One of the goals of Bayesian inference is to provide uncertainty estimates by using the \emph{posterior distribution} obtained using Bayes' rule. Unfortunately, this is infeasible in large models such as Bayesian neural networks.
Traditional methods such as Markov Chain Monte Carlo (MCMC) methods converge slowly and might require a large memory \cite{balan2015bayesian}.
In contrast, variational inference (VI) methods can scale to large models by using stochastic-gradient (SG) methods, as recent work has shown \citep{ graves2011practical, blundell2015weight, ranganath2013black, salimans2013fixed}. These works employ adaptive learning-rate methods, such as RMSprop \citep{hintonTieleman}, Adam \citep{kingma2014adam} and AdaGrad \citep{duchi2011adaptive}, for which easy-to-use implementations are available in existing codebases.

Despite their simplicity, these VI methods require more computation, memory, and implementation effort compared to maximum-likelihood estimation (MLE). 
One reason for this is that the number of parameters in VI is usually much larger than in MLE, which increases the memory and computation costs.
Another reason is that existing codebases are designed and optimized for tasks such as MLE, and their application to VI involves significant amount of modifications in the code. 
We ask the following question: is it possible to avoid these issues and make VI as easy as MLE? 

In this paper, we propose to use \emph{natural-gradient methods} to address these issues for Gaussian mean-field VI. By proposing a \emph{natural-momentum method} along with a series of approximations, we obtain algorithms that can be implemented with minimal changes to the existing codebases of adaptive learning-rate methods.
The main change involves perturbing the network weights during the gradient computation (see Fig.~\ref{fig:adamVsVadam}).
An uncertainty estimate can be cheaply obtained by using the vector that adapts the learning rate. 
This requires lower memory, computation, and implementation efforts than existing methods for VI while obtaining uncertainty estimates of comparable quality.
Our experimental results confirm this, and suggest that the estimated uncertainty could improve exploration in problems such as reinforcement learning and stochastic optimization.

\begin{figure*}[!t]
	\center
	\fbox{
		\subfigure{
			\begin{minipage}{.42\textwidth}
				\textbf{Adam}
				\begin{algorithmic}[1]
					\WHILE{not converged}
					\STATE $\vtheta \leftarrow \vmu$ 
               \STATE Randomly sample a data example $\data_i$
               \STATE $\vg \leftarrow - \nabla \log p(\data_i|\vtheta) $
					\STATE $\vm \leftarrow \gamma_1 \, \vm + (1-\gamma_1) \, \vg $ 
               \STATE $\vs \leftarrow \gamma_2 \, \vs + (1-\gamma_2) \, (\vg \circ \vg)$ 
					\STATE $\hat{\vm} \leftarrow \vm/(1-\gamma_1^t), \quad \hat{\vs} \leftarrow \vs/(1-\gamma_2^t)$
					\STATE $\vmu \leftarrow \vmu - \alpha \,\, \hat{\vm} / (\sqrt{\hat{\vs}} +  \delta )$
					\STATE $t \leftarrow t + 1$
					\ENDWHILE
				\end{algorithmic}
	\end{minipage}}}
   \hfill
	\fbox{%
		\subfigure{
			\begin{minipage}{.52\textwidth}
				\textbf{Vadam}
				\begin{algorithmic}[1]
					\WHILE{not converged}
            \STATE $\vtheta \leftarrow \vmu {\color{red}\, +\, \vsigma\circ \vepsilon}$, where $\vepsilon \sim \gauss(0,\vI)$, $\vsigma\leftarrow 1/\sqrt{N\vs + \lambda}$
               \STATE Randomly sample a data example $\data_i$
               \STATE $\vg \leftarrow - \nabla \log p(\data_i|\vtheta) $
					\STATE $\vm \leftarrow \gamma_1 \, \vm + (1-\gamma_1) \, \rnd{\vg {\color{red} \,\, + \,\, \lambda \vmu/N } }$ 
               \STATE $\vs \leftarrow \gamma_2 \, \vs + (1-\gamma_2) \, (\vg \circ \vg)$ 
					\STATE $\hat{\vm} \leftarrow \vm/(1-\gamma_1^t), \quad \hat{\vs} \leftarrow \vs/(1-\gamma_2^t)$
               \STATE $\vmu \leftarrow \vmu - \alpha \,\, \hat{\vm} / (\sqrt{\hat{\vs}} + {\color{red} \lambda/N})$
					\STATE $t \leftarrow t + 1$
					\ENDWHILE
				\end{algorithmic}
	\end{minipage}}}
   \caption{Comparison of Adam (left) and one of our proposed method Vadam (right). Adam performs maximum-likelihood estimation while Vadam performs variational inference, yet the two pseudocodes differ only slightly (differences highlighted in red). A major difference is in line 2 where, in Vadam, weights are perturbed during the gradient evaluations.}
	\label{fig:adamVsVadam}
\end{figure*}

\subsection{Related Work}
Bayesian inference in models such as neural networks has a long history in machine learning \citep{mackay2003information,bishop2006pattern}. Earlier work proposed a variety of algorithms such as MCMC methods \citep{neal95}, Laplace's method \citep{denker1991transforming}, and variational inference \citep{hinton1993keeping,barber1998ensemble}.
The mean-field approximation has also been a popular tool from very early on \citep{saul1996mean, anderson1987mean}.
These previous works lay the foundation of methods now used for Bayesian deep learning \citep{Gal2016Uncertainty}. 

Recent approaches \citep{graves2011practical, blundell2015weight} enable the application of Gaussian mean-field VI methods to large deep-learning problems. They do so by using gradient-based methods. In contrast, we propose to use \emph{natural}-gradient methods which, as we show, lead to algorithms that are simpler to implement and require lower memory and computations than gradient-based methods.
Natural gradients are also better suited for VI because they can improve convergence rates by exploiting the information geometry of posterior approximations \cite{khan2016faster}. Some of our algorithms inherit these properties too.

A recent independent work on noisy-Adam by~\citet{DBLP:journals/corr/abs-1712-02390} is algorithmically very similar to our Vadam method, however their derivation lacks a strong motivation for the use of momentum. In our derivation, we incorporate a \emph{natural-momentum} term based on Polyak's heavy-ball method, which provides a theoretical justification for the use of momentum. In addition, we analyze the approximation error introduced in Vadam and discuss ways to reduce it.

\citet{DBLP:journals/corr/abs-1712-02390} also propose an interesting extension by using K-FAC, which could find better approximations than the mean-field method. The goal of this approach is similar to other approaches that employ \emph{structured} approximations \citep{ritter2018scalable, louizos2016structured, sun2017learning}.
Many other works have explored variety of approximation methods, e.g., \citet{yarin16dropout} use dropout for VI, \citet{hernandez15pbp, hasenclever2017distributed} use expectation propagation, \citet{li2016preconditioned, balan2015bayesian} use stochastic-gradient Langevin dynamics. Such approaches are viable alternatives to the mean-field VI approach we use.

Another related work by \citet{mandt2017stochastic} views SG descent as VI but requires additional effort to obtain posterior approximations, while in our approach the approximation is automatically obtained within an adaptive method. 

Our weight-perturbed algorithms are also related to global-optimization methods, e.g., Gaussian-homotopy continuation methods \citep{mobahi2015theoretical}, smoothed-optimization method \citep{leordeanu2008smoothing}, graduated optimization method \citep{hazan2016graduated}, and stochastic search methods \citep{zhou2014gradient}.
In particular, our algorithm is related to recent approaches in deep learning for exploration to avoid local minima, e.g., natural evolution strategy \cite{wierstra2008natural}, entropy-SGD \cite{chaudhari2016entropy}, and noisy networks for reinforcement learning \cite{fortunato2017noisy, plappert2017parameter}.  
An earlier version of our work \cite{2017arXiv171105560E} focuses exclusively on this problem, and in this paper we modify it to be implemented within an adaptive algorithm like Adam.

\section{Gaussian Mean-Field Variational Inference}
\label{sec:gaussVI}
We consider modeling of a dataset $\data = \{ \data_1,\data_2,\ldots,\data_N \}$ by using a deep neural network (DNN). We assume a probabilistic framework where each data example $\data_i$ is sampled independently from a probability distribution $p(\data_i|\vtheta)$ parameterized by a DNN with weights $\vtheta\in\real^D$, e.g., the distribution could be an exponential-family distribution whose mean parameter is the output of a DNN
\cite{bishop2006pattern}.

One of the most popular approaches to estimate $\vtheta$ given $\data$ is maximum-likelihood estimation (MLE), where we maximize the log-likelihood: $\log p(\data|\vtheta)$.
This optimization problem can be efficiently solved by applying SG methods such as RMSProp, AdaGrad and Adam.	
For large problems, these methods are extremely popular, partly due to the simplicity and efficiency of their implementations (see Fig.~\ref{fig:adamVsVadam} for Adam's pseudocode).

One of the goals of Bayesian deep learning is to go beyond MLE and estimate the \emph{posterior distribution} of $\vtheta$ to obtain an uncertainty estimate of the weights.
Unfortunately, the computation of the posterior is challenging in deep models.
The posterior is obtained by specifying a \emph{prior distribution} $p(\vtheta)$ and then using Bayes' rule: $p(\vtheta| \data) := p(\data|\vtheta) p(\vtheta)/p(\data)$.
This requires computation of the normalization constant $p(\data) = \int p(\data|\vtheta) p(\vtheta) d\vtheta$ which is a very difficult task for DNNs.
One source of the difficulty is the size of $\vtheta$ and $\data$ which are usually very large in deep learning.
Another source is the \emph{nonconjugacy} of the likelihood $p(\data_i|\vtheta)$ and the prior $p(\vtheta)$, i.e., the two distributions do not take the same form with respect to $\vtheta$ \cite{bishop2006pattern}. 
As a result, the product $p(\data|\vtheta) p(\vtheta)$ does not take a form with which $p(\data)$ can be easily computed.
Due to these issues, Bayesian inference in deep learning is computationally challenging.
 
Variational inference (VI) simplifies the problem by approximating $p(\vtheta| \data)$ with a distribution $q(\vtheta)$ whose normalizing constant is relatively easier to compute. 
Following previous work \cite{ranganath2013black, blundell2015weight, graves2011practical}, we choose both $p(\vtheta)$ and $q(\vtheta)$ to be Gaussian distributions with diagonal covariances:
\begin{align}
p(\vtheta):=\gauss(\vtheta| 0, \vI/\lambda), \quad 
q(\vtheta):= \gauss(\vtheta|\vmu,\diag(\vsigma^2)),
\end{align}
where $\lambda\in\real$ is a known precision parameter with $\lambda>0$, and $\vmu,\vsigma\in\real^D$ are mean and standard deviation of $q$. The distribution $q(\vtheta)$ is known as the \emph{Gaussian mean-field variational distribution} and its parameters $\vmu$ and $\vsigma^2$ can be obtained by maximizing the following \emph{variational objective}:
\begin{align}
   \mathcal{L}(\vmu,\vsigma^2) := \sum_{i=1}^N \myexpect_q \sqr{\log p(\data_i|\vtheta)} + \myexpect_{q}\sqr{\log \frac{p(\vtheta)}{ q(\vtheta)}} . \label{eq:elbo}
\end{align}
A straightforward approach used in the previous work \cite{ranganath2013black, blundell2015weight, graves2011practical} is to maximize $\mathcal{L}$ by using an SG method, e.g., we can use the following update:
\begin{align}
   \vmu_{t+1} = \vmu_t + \rho_t \sg{\mathcal{L}_t}{\mu} , \quad \quad 
\vsigma_{t+1} = \vsigma_t + \delta_t \sg{\mathcal{L}_t}{\sigma}  , \label{eq:vsgd_cov}
\end{align}
where $t$ is the iteration number, $\widehat{\nabla}_x \mathcal{L}_t$ denotes an unbiased SG estimate of $\mathcal{L}$ at $\vmu_t,\vsigma_t^2$ with respect to $x$, and $\rho_t,\delta_t>0$ are learning rates which can be adapted using methods such as RMSprop or AdaGrad.
These approaches make use of existing codebases for adaptive learning-rate methods to perform VI, which can handle many network architectures and can scale well to large datasets.

Despite this, a direct application of adaptive learning-rate methods for VI may result in algorithms that use more computation and memory than necessary, and also require more implementation effort.
Compared to MLE, the memory and computation costs increase because the number of parameters to be optimized is doubled and we now have two vectors $\vmu$ and $\vsigma$ to estimate.
Using adaptive methods increases this cost further as these methods require storing the scaling vectors that adapt the learning rate for both $\vmu$ and $\vsigma$.
In addition, using existing codebases require several modifications as they are designed and optimized for MLE.
For example, we need to make changes in the computation graph where the objective function is changed to the variational objective and network weights are replaced by random variables.
Together, these small issues make VI more difficult to implement and execute than MLE. 

The algorithms developed in this paper solve some of these issues and can be implemented within Adam with minimal changes to the code. We derive our algorithm by approximating a natural-gradient method and then using a natural-momentum method. We now describe our method in detail.

\section{Approximate Natural-Gradient VI}
\label{sec:vprop}
In this section, we introduce a natural-gradient method to perform VI and then propose several approximations that enable implementation within Adam.

Natural-gradient VI methods exploit the Riemannian geometry of $q(\vtheta)$ by scaling the gradient with the inverse of its Fisher information matrix (FIM).
We build upon the natural-gradient method of \citet{khan2017conjugate}, which simplifies the update by avoiding a direct computation of the FIM. The main idea is to use the \emph{expectation parameters of the exponential-family distribution to compute natural gradients in the natural-parameter space}. We provide a brief description of their method in Appendix \ref{app:review}.

For Gaussian mean-field VI, the method of \citet{khan2017conjugate} gives the following update: 
\begin{align}
   \textrm{NGVI : } \vmu_{t+1} &= \vmu_{t} + \beta_t \,\, \vsigma^2_{t+1} \circ \sqr{\widehat{\nabla}_\mu {\mathcal{L}_t}},  \label{eq:Van_mean_0}\\
   \vsigma_{t+1}^{-2} &= \vsigma_t^{-2} - \,\, 2\beta_t \,\, \sqr{ \widehat{\nabla}_{\sigma^2} \mathcal{L}_t},  \label{eq:Van__0}
\end{align}
where $\beta_t > 0$ is a scalar learning rate and $\va\circ\vb$ denotes the element-wise product between vectors $\va$ and $\vb$.~We refer to this update as natural-gradient variational inference (NGVI). A detailed derivation is given in Appendix \ref{app:mirror}.

The NGVI update differs from \eqref{eq:vsgd_cov} in one major aspect: the learning rate $\beta_t$ in \eqref{eq:Van_mean_0} is adapted by the variance $\vsigma_{t+1}^2$. 
This plays a crucial role in reducing the NGVI update to an Adam-like update, as we show the next section. The update requires a constraint $\vsigma^2>0$ but, as we show in Section \ref{sec:vogn}, we can eliminate this constraint using an approximation.

\subsection{Variational Online-Newton (VON)}
We start by expressing the NGVI update in terms of the MLE objective, so that we can directly compute gradients on the MLE objective using backpropagation.
We start by defining the MLE objective (denoted by $f$) and minibatch stochastic-gradient estimates (denoted by $\hat{\vg}$):
\begin{align}
   f(\vtheta) := \frac{1}{N} \sum_{i=1}^N f_i(\vtheta), \quad 
   \hat{\vg}(\vtheta) := \frac{1}{M} \sum_{i\in\minibatch} \nabla_\theta f_i(\vtheta), \label{eq:mle_obj}
\end{align}
where $f_i(\vtheta) := -\log p(\data_i|\vtheta)$ is the negative log-likelihood of $i$'th data example, and the minibatch $\minibatch$ contains $M$ examples chosen uniformly at random. Similarly, we can obtain a minibatch stochastic-approximation of the Hessian which we denote by $\widehat{\nabla}_{\theta\theta}^2 f(\vtheta)$.

As we show in Appendix \ref{app:von}, the NGVI update can be written in terms of the stochastic gradients and Hessian of $f$:
\begin{align}
   \textrm{VON : }&\vmu_{t+1} = \vmu_{t} - \beta_t\,\, ( \hat{\vg}(\vtheta_t) + \tlambda\vmu_t)/(\vs_{t+1} + \tlambda) ,  \label{eq:Von_mu_0}\\
                       &\vs_{t+1} = (1-\beta_t) \vs_t +  \beta_t \,\, \diag[ \widehat{\nabla}_{\theta\theta}^2 f(\vtheta_t)], 
\label{eq:Von_prec_0}
\end{align}
where $\va/\vb$ is an element-wise division operation between vectors $\va$ and $\vb$, and we have approximated the expectation with respect to $q$ using one Monte-Carlo (MC) sample $\vtheta_t \sim \gauss(\vtheta|\vmu_t,\vsigma_t^2)$ with $\vsigma^2_t := 1/[N(\vs_t + \tlambda)]$ and $\tlambda := \lambda/N$. The update can be easily modified when multiple samples are used. This update can leverage backpropagation to perform the gradient and Hessian computation. Since the scaling vector $\vs_t$ contains an online estimate of the diagonal of the Hessian, we call this the
``variational online-Newton" (VON) method. VON is expected to perform as well as NGVI, but does not require the gradients of the variational objective. 

The Hessian can be computed by using methods such as automatic-differentiation or the reparameterization trick. However, since $f$ is a non-convex function, the Hessian can be negative which might make $\vsigma^2$ negative, in which case the method will break down. One could use a constrained optimization method to solve this issue, but this might be difficult to implement and execute (we discuss this briefly in Appendix \ref{app:hess_reparam}). In the next section, we propose a simple fix
to this problem by using an approximation.

\subsection{Variational Online Gauss-Newton (VOGN)}
\label{sec:vogn}
To avoid negative variances in the VON update, we propose to use the Generalized Gauss-Newton (GGN) approximation \citep{schraudolph2002fast,martens2014new, graves2011practical}:
\begin{align}
   \nabla_{\theta_j\theta_j}^2 f(\vtheta) \approx \frac{1}{M} \sum_{i \in \minibatch} \sqr{ \nabla_{\theta_j} f_i(\theta) }^2 := \hat{h}_j(\vtheta), \label{eq:ggn}
\end{align}
where $\theta_j$ is the $j$'th element of $\vtheta$. This approximation will always be nonnegative, therefore if the initial $\vsigma^2$ at $t=1$ is positive, it will remain positive in the subsequent iterations. Using this approximation to update $\vs_t$ in \eqref{eq:Von_prec_0} and denoting the vector of $\hat{h}_j(\vtheta)$ by $\hat{\vh}(\vtheta)$, we get,
\begin{align}
   \textrm{VOGN : }\,\, & \vs_{t+1} = (1-\beta_t) \vs_t +  \beta_t \,\, \hat{\vh}(\vtheta_t).
\label{eq:VOGN-D}
\end{align}
Using this update in VON, we get the ``variational online Gauss-Newton'' (VOGN) algorithm.

The GGN approximation is proposed by \citet{graves2011practical} for mean-field Gaussian VI to derive a fast gradient-based method (see Eq. (17)~in his paper\footnote{There is a discrepancy between Eq. (17) and (12) in \citet{graves2011practical}, however the text below Eq. (12) mentions relationship to FIM, from which it is clear that the GGN approximation is used.}). This approximation is very useful for our natural-gradient method since it eliminates the constraint on $\vsigma^2$, giving VOGN an algorithmic advantage over VON.

How good is this approximation? For an MLE problem, the approximation error of the GGN in~\eqref{eq:ggn} decreases as the model-fit improves during training \citep{martens2014new}. 
For VI, we expect the same however, since $\vtheta$ are sampled from $q$, the expectation of the error is unlikely to be zero. Therefore, the solutions found by VOGN will typically differ from those found by VON, but their performances are expected to be similar.

An issue with VOGN is that its implementation is not easy within existing deep-learning codebases. This is because these codebases are optimized to directly compute the sum of the gradients over minibatches, and do not support computation of individual gradients as required in~\eqref{eq:ggn}. 
A solution for such computations is discussed by \citet{goodfellow2015efficient}, but this requires additional implementation effort.
Instead, we address this issue by using another approximation in the next section.

\subsection{Variational RMSprop (Vprop)}
\label{subsec:vprop}
To simplify the implementation of VOGN, we propose to approximate the Hessian by the \emph{gradient magnitude} (GM) \citep{bottou2016optimization}:
\begin{align}
   \nabla_{\theta_j \theta_j}^2 f(\vtheta) &\approx \sqr{ \frac{1}{M} \sum_{i \in \minibatch}  \nabla_{\theta_j} f_i(\theta) }^2 = \sqr{\hat{g}_j(\vtheta)}^2 . \label{eq:gm}
\end{align}
Compared to the GGN which computes the sum of squared-gradients, this approximation instead computes the square of the sum. This approximation is also used in RMSprop which uses the following update given weights $\vtheta_t$:
\begin{align}
   \textrm{RMSprop} &:  \vtheta_{t+1} = \vtheta_{t} - \alpha_t \,\, \hat{\vg}(\vtheta_t)  /(\sqrt{\bar{\vs}_{t+1}} + \delta) ,
\label{eq:rmsprop} \\
   &\bar{\vs}_{t+1} = (1-\beta_t) \bar{\vs}_t +  \beta_t  \sqr{ \hat{\vg}(\vtheta_t)\circ\hat{\vg}(\vtheta_t)},
\label{eq:rmsprop_s}
\end{align}
where $\bar{\vs}_t$ is the vector that adapts the learning rate and $\delta$ is a small positive scalar added to avoid dividing by zero.
The update of $\bar{\vs}_t$ uses the GM approximation to the Hessian \citep{bottou2016optimization}. Adam and AdaGrad also use this approximation.

Using the GM approximation and an additional modification in the VON update, we can make the VON update very similar to RMSprop. Our modification involves taking the square-root over $\vs_{t+1}$ in \eqref{eq:Von_mu_0} and then using the GM approximation for the Hessian. We also use different learning rates $\alpha_t$ and $\beta_t$ to update $\vmu$ and $\vs$, respectively. The resulting update is very similar to the RMSprop update:
\begin{align}
   \textrm{Vprop: }\,\, &\vmu_{t+1} = \vmu_{t} - \alpha_t\,\, ( \hat{\vg}(\vtheta_t) {\color{red} + \tlambda\vmu_t})/(\sqrt{\vs_{t+1}} + {\color{red} \tlambda}) ,  \nonumber \\
                  &\vs_{t+1} = (1-\beta_t) \vs_t +  \beta_t \,\, \sqr{ \hat{\vg}(\vtheta_t)\circ\hat{\vg}(\vtheta_t)} , 
\label{eq:vprop_prec_0}
\end{align}
where \textcolor{red}{$\vtheta_t \sim \gauss(\vtheta|\vmu_t,\vsigma_t^2)$} with $\vsigma^2_t := 1/[N(\vs_t + \tlambda)]$.  We call this update ``Variational RMSprop" or simply ``Vprop".

The Vprop update resembles RMSprop but with three differences (highlighted in red).
First, the gradient in Vprop is evaluated at the weights $\vtheta_t$ sampled from  $\mathcal{N}(\vtheta | \vmu_t, \vsigma_t^2)$. This is a \emph{weight-perturbation} where the variance $\vsigma_t^2$ of the perturbation is obtained from the vector $\vs_t$ that adapts the learning rate. The variance is also the uncertainty estimates. Therefore, VI can be performed simply by using an RMSprop update with a few simple changes. The second difference between Vprop and RMSprop is that Vprop has an extra term $\tlambda \vmu_t$ in
the update of $\vmu_t$ which is due to the Gaussian prior. Finally, the third difference is that the constant $\delta$ in RMSprop is replaced by $\tlambda$.

\subsection{Analysis of the GM approximation}
\label{sec:analysis}

It is clear that the GM approximation might not be the best approximation of the Hessian. Taking square of a sum leads to a sum with $M^2$ terms which, depending on the correlations between the individual gradients, would either shrink or expand the estimate. The following theorem formalizes this intuition. It states that, given a minibatch of size $M$, the expectation of the GM approximation is somewhere between the GGN and square of the full-batch gradient.
\begin{thm}
   \label{thm:11}
   Denote the full-batch gradient with respect to $\theta_j$ by $g_j(\vtheta)$ and the corresponding full-batch GGN approximation by $h_j(\vtheta)$. Suppose minibatches $\minibatch$ are sampled from the uniform distribution $p(\minibatch)$ over all ${N \choose M}$ minibatches, and denote a minibatch gradient by $\hat{g}_j(\vtheta; \minibatch)$, then the expected value of the GM approximation is the following,
\begin{align}
   \myexpect_{p(\minibatch)} \sqr{\hat{g}_j(\vtheta; \minibatch)^2} = w h_j(\vtheta) + (1-w) [g_j(\vtheta)]^2,
\end{align}
where $w = \frac{1}{M}(N-M)/(N-1)$. 
\end{thm}
A proof is given in Appendix \ref{app:proof1}. This result clearly shows the bias introduced in the GM approximation and also that the bias increases with the minibatch size. For a minibatch of size $M=1$, we have $w=1$ and the GM is an unbiased estimator of the GGN, but when $M=N$ it is purely the magnitude of the gradient and does not contain any second-order information.

Therefore, if our focus is to obtain uncertainty estimates with good accuracy, VOGN with $M=1$ might be a good choice since it is as easy as Vprop to implement. However, this might require a small learning-rate and converge slowly.
Vprop with $M>1$ will converge fast and is much easier to implement than VOGN with $M>1$, but might result in slightly worse estimates.
Using Vprop with $M=1$ may not be as good because of the square-root\footnote{Note that the square-root does not affect a fixed point (see Appendix \ref{app:fixed}) but it might still affect the steps taken by the algorithm.} over $\vs_t$.

\section{Variational Adam (Vadam)}
\label{sec:vadam}
We now propose a \emph{natural-momentum} method which will enable an Adam-like update.	
	
	Momentum methods generally take the following form that uses Polyak's heavy-ball method:
	\begin{align}
	\vtheta_{t+1} = \vtheta_t + \bar{\alpha}_t \nabla_\theta f_1(\vtheta_t) + \bar{\gamma}_t (\vtheta_t -\vtheta_{t-1}),
	\label{eq:momen}
	\end{align}
   where $f_1$ is the function we want to maximize and the last term is the momentum term. We propose a \emph{natural-momentum} version of this algorithm which employs natural-gradients instead of the gradients. We assume $q$ to be an exponential-family distribution with natural-parameter $\veta$. We propose the following natural-momentum method in the natural-parameter space: 
   \begin{align}
      \veta_{t+1} = \veta_t + \bar{\alpha}_t \widetilde{\nabla}_\eta \mathcal{L}_t + \bar{\gamma}_t (\veta_t -\veta_{t-1}),
	\label{eq:nat_momen}
	\end{align}
   where $\widetilde{\nabla}$ denotes the natural-gradients in the natural-parameter space, i.e., the gradient scaled by the Fisher information matrix of $q(\vtheta)$.

   We show in Appendix \ref{app:vadam_deriv} that, for Gaussian $q(\vtheta)$, we can express the above update as a VON update with momentum\footnote{This is not an exact update for \eqref{eq:nat_momen}. We make one approximation in Appendix \ref{app:vadam_deriv} to derive it.}:
\begin{align}
   \vmu_{t+1} &= \vmu_{t} - \bar{\alpha}_t \,\, \sqr{ \frac{1}{\vs_{t+1}  + \tlambda} } \circ \rnd{ \nabla_\theta            f(\vtheta_t) + \tlambda\vmu_t}  \nonumber \\
        &\quad\quad\quad\quad\quad + \bar{\gamma}_t \sqr{ \frac{\vs_t + \tlambda}{\vs_{t+1} + \tlambda} }\circ (\vmu_t - \vmu_{t-1}), \label{eq:von_momen21}\\
   \vs_{t+1} &= \rnd{1-\bar{\alpha}_t} \vs_t +  \bar{\alpha}_t \,\,  \nabla_{\theta\theta}^2 f(\vtheta_t), \label{eq:von_momen22}
 \end{align}
 where $\vtheta_t \sim \gauss(\vtheta|\vmu_t,\vsigma_t^2)$ with $\vsigma_t^2 = 1/[N(\vs_t + \tlambda)]$.
 This update is similar to \eqref{eq:nat_momen}, but here the learning rates are adapted. An attractive feature of this update is that it is very similar to Adam.
 Specifically the Adam update shown in Fig. \ref{fig:adamVsVadam} can be expressed as the following adaptive version of \eqref{eq:momen} as shown in \citet{wilson2017marginal}\footnote{\citet{wilson2017marginal} do not use the constant $\delta$, but in Adam a small constant is added for numerical stability.},
\begin{align}
\vtheta_{t+1} &= \vtheta_t - \tilde{\alpha}_t \sqr{ \frac{1}{\sqrt{\hat{\vs}_{t+1}}  + \delta} } \circ \nabla_\theta f(\vtheta_t) \nonumber \\
				&\quad\quad\quad\quad+ \tilde{\gamma}_t \sqr{ \frac{\sqrt{\hat{\vs}_t} + \delta }{\sqrt{\hat{\vs}_{t+1}} + \delta } } \circ (\vtheta_t - \vtheta_{t-1}) ,
\label{eq:adam_scaled_momen1} \\
\hat{\vs}_{t+1} &= \gamma_2 \hat{\vs}_t +  (1-\gamma_2) \,\,  [\hat{\vg}(\vtheta_t)]^2,
\end{align}
where $\tilde{\alpha}_t, \tilde{\gamma}_t$ are appropriately defined in terms of the Adam's learning rate $\alpha$ and $\gamma_1$: $\tilde{\alpha}_t := \alpha \rnd{1-\gamma_1}/\rnd{1-\gamma_1^t}$ and $\tilde{\gamma}_t := \gamma_1\rnd{1-\gamma_1^{t-1}}\rnd{1-\gamma_1^t}$.

Using a similar procedure as the derivation of Vprop, we can express the update as an Adam-like update, which we call ``variational Adam" or simply ``Vadam". A pseudocode is given in Fig. \ref{fig:adamVsVadam}, where we use learning rates of the Adam update insteof of choosing them accoring to $\bar{\alpha}_t$ and $\bar{\gamma}_t$. A derivation is given in Appendix \ref{eq:von_momen}.

\section{Variational AdaGrad (VadaGrad)}
\label{sec:vadagrad}
Vprop and Vadam perform variational inference, but they can be modified to perform optimization instead of inference. We now derive such an algorithm which turns out to be a variational version of AdaGrad.

We follow \citet{2012arXiv12124507S} who consider minimization of black-box functions $F(\vtheta)$ via the \emph{variational optimization}\footnote{The exact conditions on $F$ under which VO can be applied are also discussed by \citet{2012arXiv12124507S}.} (VO) framework. In this framework, instead of directly minimizing $F(\vtheta)$, we minimize its expectation $\myexpect_{q} \sqr{F(\vtheta)}$ under a distribution $q(\vtheta) := \gauss(\vtheta|\vmu,\vsigma^2)$ with respect to $\vmu$ and $\vsigma^2$. 
The main idea behind VO is that the expectation can be used as a surrogate to the original optimization problem since $\min_{\theta} F(\vtheta) \le \myexpect_{q} \sqr{F(\vtheta)}$.
The equality is attained when $\vsigma^2 \to 0$, i.e., all mass of $\mathcal{N}(\vtheta|\vmu,\vsigma^2)$ is at the mode.
	The main advantage of VO is that $\myexpect_{q} \sqr{F(\vtheta)}$ is differentiable even when $F$ itself is non-differentiable.
	This way we can use SG optimizers to solve such problems.

   Similarly to Vprop, we can derive an algorithm for VO by noting that VO can be seen as a special case of the VI problem \eqref{eq:elbo} where the KL term is absent and $F(\vtheta)$ is the negative log-likelihood. With this in mind, we define the following variational objective with an additional parameter $\tau \in [0,1]$:
	\begin{align}
      \mathcal{L}_F(\vmu,\vsigma^2) :=  -\myexpect_q \sqr{F(\vtheta)} + \tau \myexpect_{q}\sqr{\log \frac{p(\vtheta)}{ q(\vtheta)}} . \label{eq:elbo_tau}
	\end{align}
	The parameter $\tau$ allows us to interpolate between inference and optimization. 
   When $\tau=1$, the objective corresponds to VI with a negative log-likelihood $F(\vtheta)$, and when $\tau=0$, it corresponds to VO. 
    Similar objectives have been proposed in existing works \cite{blundell2015weight, higgins2016beta} where $\tau$ is used to improve convergence.
	
    For twice-differentiable $F$, we can follow a similar derivation as Section~\ref{sec:vprop}, and obtain the following algorithm,
\begin{align}
   &\vmu_{t+1} = \vmu_{t} - \alpha_t\,\, ( \widehat{\nabla}_{\theta} F(\vtheta) +  \tau \lambda\vmu_t)/(\vs_{t+1} + \tau\lambda) ,  \label{eq:mu123}\\
   &\vs_{t+1} = (1- \tau \beta_t) \vs_t + \beta_t \,\,  \widehat{\nabla}_{\theta\theta}^2 F(\vtheta), 
\label{eq:prec123}
\end{align}
where $\vtheta_t \sim \gauss(\vtheta|\vmu_t,\vsigma_t^2)$ with $\vsigma^2_t := 1/(\vs_t + \tau\lambda)$. This algorithm is identical to the VON algorithm when $\tau=1$, but when $\tau=0$, we perform VO with an algorithm which is a diagonal version of the  Variational Adaptive-Newton (VAN) algorithm proposed in \citet{2017arXiv171105560E}. By setting the value of $\tau$ between 0 and 1, we can interpolate between VO and VI. When the function is not differentiable, we can still compute the derivative of $\myexpect_q[F(\vtheta)]$ by using methods such as REINFORCE \cite{williams1992simple}.

When Hessian is difficult to compute, we can employ a GM approximation and take the square-root as we did in Vprop. For $\tau=0$, the updates turn out to be similar to AdaGrad, which we call ``variational AdaGrad" or simply ``VadaGrad". The exact updates are given in Appendix \ref{app:vadagrad}. Unlike Vprop and Vadam, the scaling vector $\vs_t$ in VadaGrad is a weighted sum of the past gradient-magnitudes. Therefore, the entries in $\vs_t$ never decrease, and the variance estimate of VadaGrad never expands. This implies that it is highly likely that
$q(\vtheta)$ will converge to a Dirac delta and therefore arrive at a minimum of $F$.

\section{Results}
\label{sec:results}
\begin{figure*}[!t]
\centering
\subfigure[]{
	\includegraphics[width=0.36\linewidth]{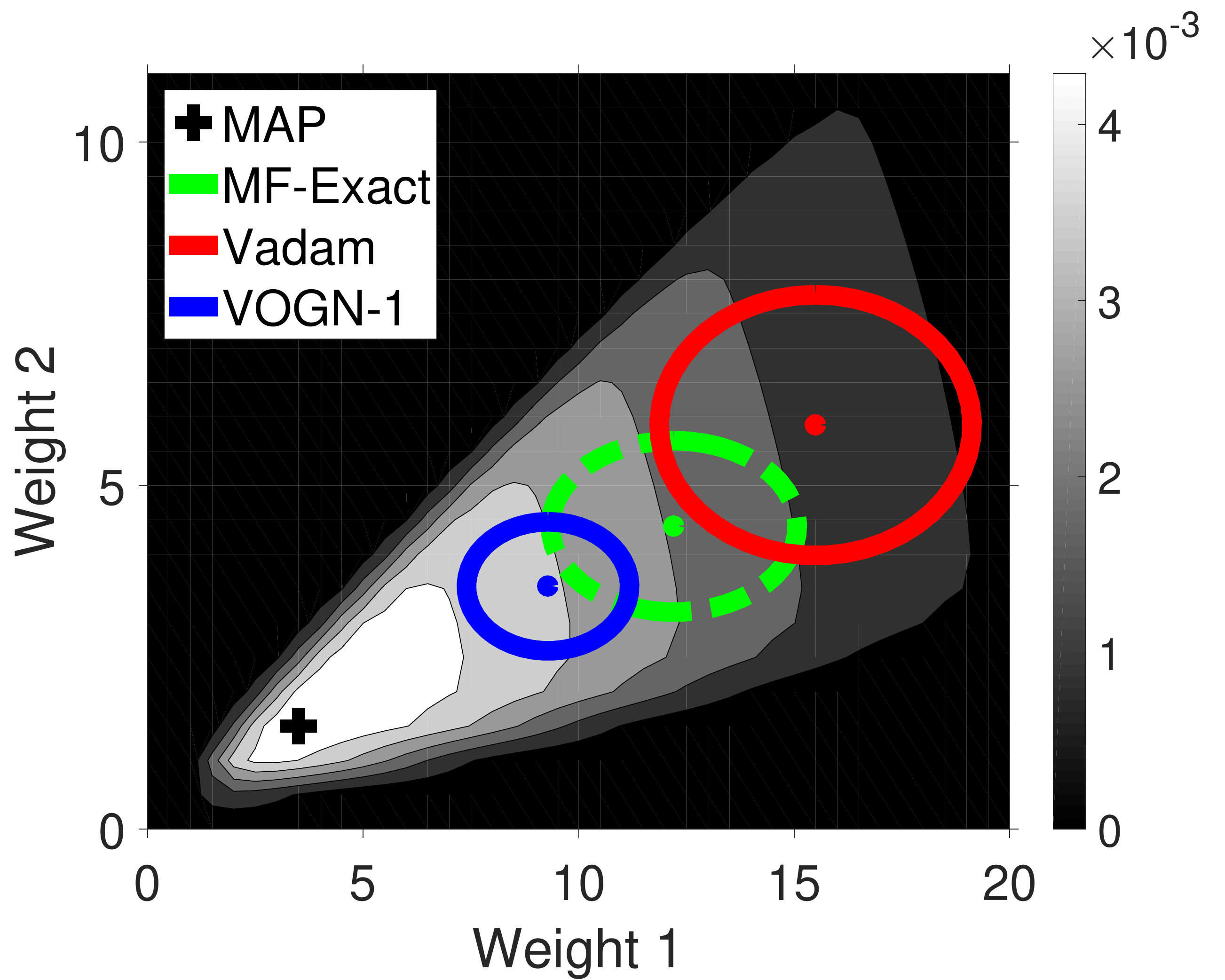}\label{figure:logreg2d}} 
\hfill
\subfigure[]{
	\includegraphics[width=0.28\linewidth]{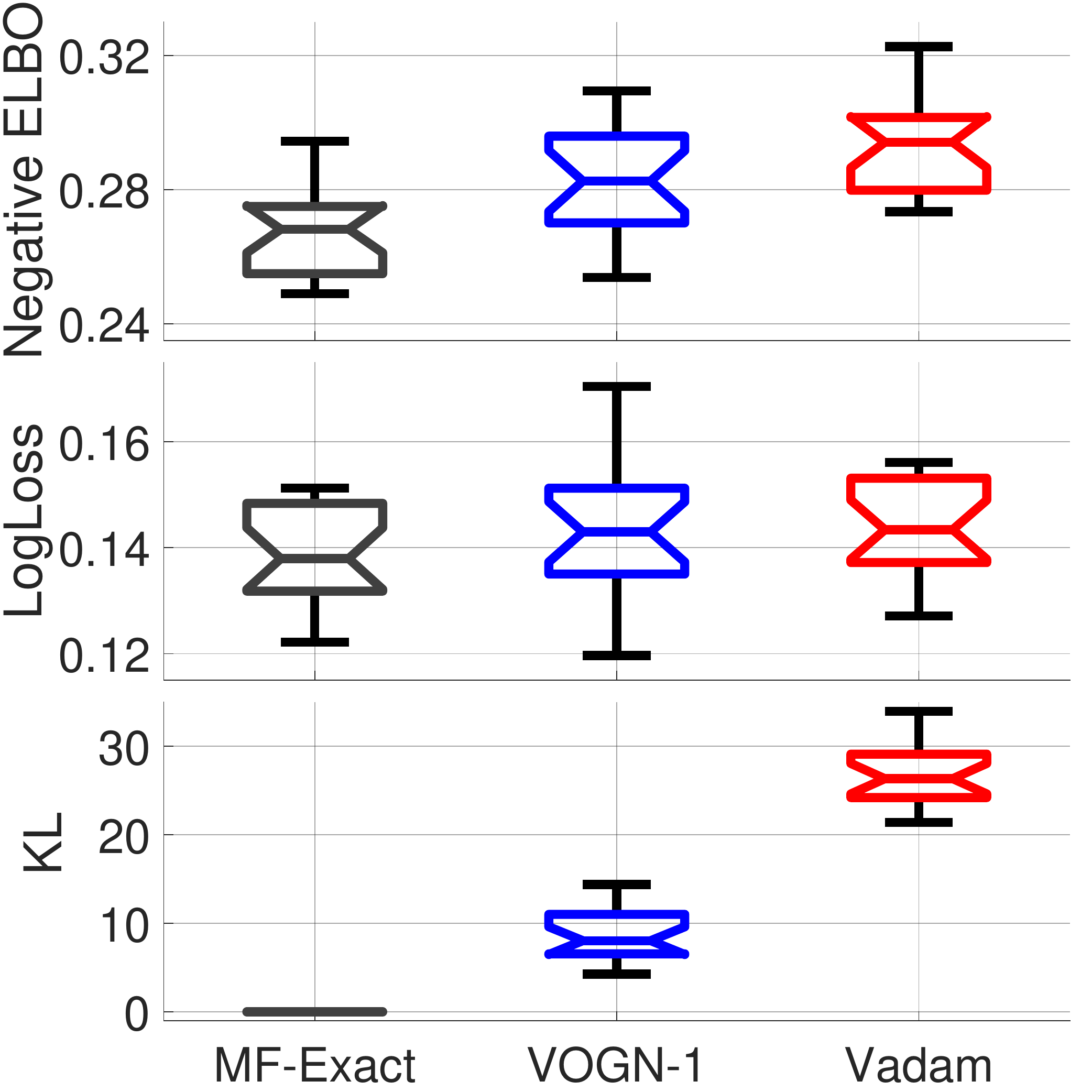}\label{figure:logregusps}}
\hfill
\subfigure[]{
	\includegraphics[width=0.29\linewidth]{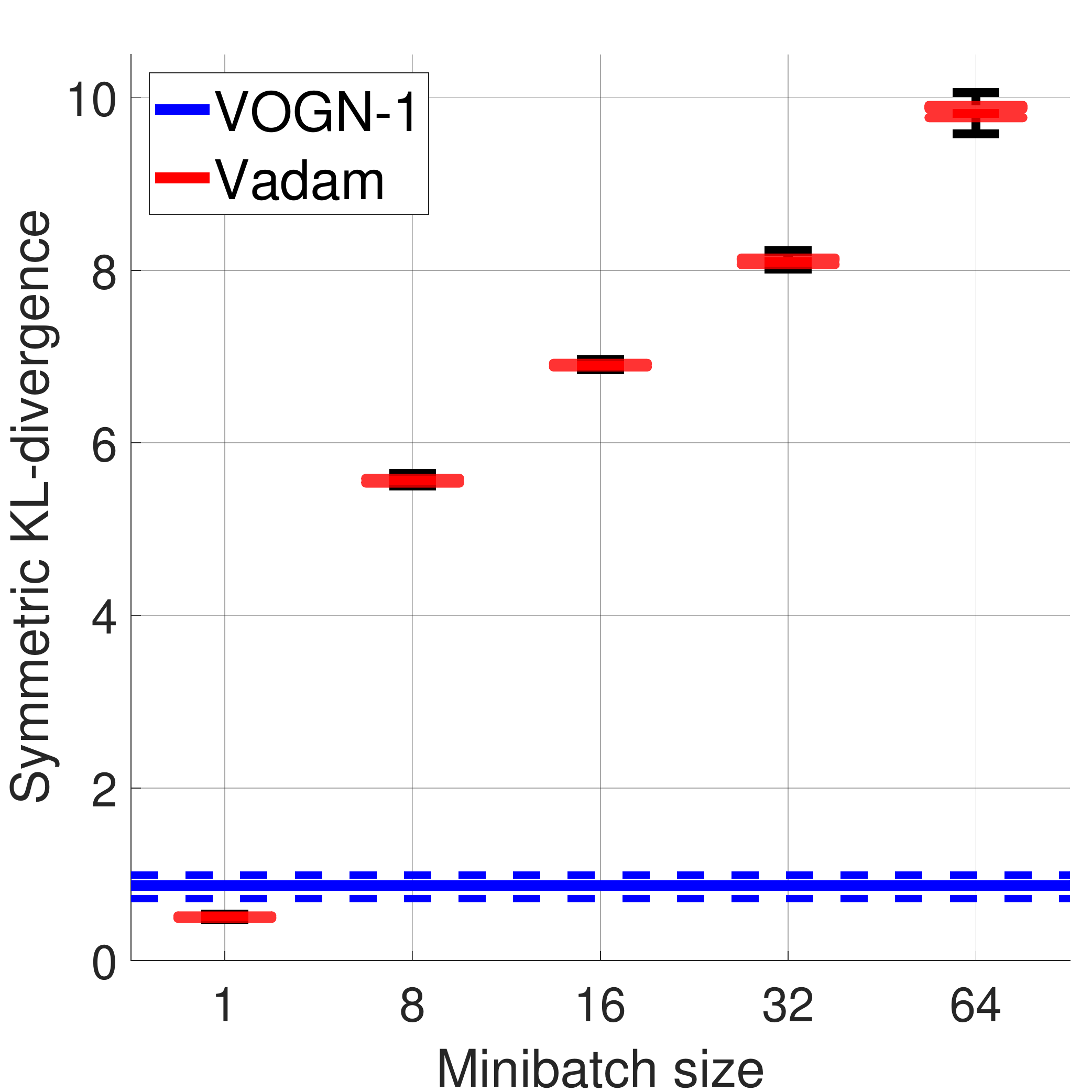}\label{figure:logregmini}}
\hfill
\caption{
Experiments on Bayesian logistic regression showing 
(a) posterior approximations on a toy example,
(b) performance on `USPS-3v5' measuring negative ELBO, log-loss, and the symmetric KL divergence of the posterior approximation to MF-Exact, 
(c) symmetric KL divergence of Vadam for various minibatch sizes on `Breast-Cancer' compared to VOGN with a minibatch of size 1. 
}
\end{figure*}

In this section, our goal is to show that the quality of the uncertainty approximations obtained using our algorithms are comparable to existing methods, and computation of uncertainty is scalable. We present results on Bayesian logistic regression for classification, Bayesian neural networks for regression, and deep reinforcement learning. An additional result illustrating avoidance of
local-minima using Vadam is in Appendix \ref{app:3d_illustration}. Another result showing benefits of weight-perturbation in Vadam is in Appendix \ref{sec:rnn}.  
The code to reproduce our results is available at {\small {\bf \url{https://github.com/emtiyaz/vadam}}}.

\subsection{Uncertainty Estimation in Logistic Regression}
\label{sec:logreg}

In this experiment, we compare the posterior approximations found with our algorithms to the optimal variational approximation that minimizes the variational objective. For Bayesian logistic regression we can compute the optimal mean-field Gaussian approximations using the method described in \citet{marlin2011piecewise} (refer to as `MF-Exact'), and compare it to the following methods: VOGN with minibatch size $M=1$ and a momentum term (referred to as
`VOGN-1'), and Vadam with $M\ge1$ (referred to as `Vadam'). Since our goal is to compare the accuracy of posterior approximations and not the speed of convergence, we run both the methods for many iterations with a small learning rate to make sure that they converge.
We use three datasets: a toy dataset ($N=60, D=2$), USPS-3vs5 ($N=1781, D=256$) and Breast-Cancer ($N=683, D=10$). Details are in Appendix \ref{app:logreg_detail}.

Fig.~\ref{figure:logreg2d} visualizes the approximations on a two-dimensional toy example from~\citet{Murphy:2012:MLP:2380985}. The true posterior distribution is shown with the contour in the background. Both, Vadam and VOGN-1 find approximations that are different from MF-Exact, which is clearly due to differences in the type of Hessian approximations they use. 

For real datasets, we compare performances using three metrics. First, the negative of the variational objective on the training data (the evidence lower-bound or ELBO), log-loss on the test data, and the symmetric KL distance between MF-Exact and the approximation found by a method. Fig. \ref{figure:logregusps} shows the results averaged over 20 random splits of the USPS-3vs5 dataset.
ELBO and log-loss are comparable for all methods, although Vadam does slightly worse on ELBO and VOGN-1 has slightly higher variance for log-loss.
However, performance on the KL distance clearly shows the difference in the quality of posterior approximations. VOGN-1 performs quite well since it uses an unbiased approximation of the GNN. Vadam does worse due to the bias introduced in the GM approximation with minibatch $M>1$, as indicated by Theorem \ref{thm:11}.

Fig. \ref{figure:logregmini} further shows the effect of $M$ where, for each $M$,  we plot results for 20 random initializations on one split of the Breast-Cancer dataset. As we decrease $M$, Vadam's performance gets better, as expected. For $M=1$, it closely matches VOGN-1. The results are still different because Vadam does not reduce to VOGN-1, even when $M=1$ due to the use of the square-root over $\vs_t$.

\begin{table*}[t]
   \setlength{\tabcolsep}{4pt}
   \centering
	\caption{Performance comparisons for BNN regression. The better method out of BBVI and Vadam is shown in boldface according to a paired t-test with $p$-value$>0.01$. Both methods perform comparably but MC-Dropout outperforms them.}
   \label{table:uci}
   \begin{tabular}{l r r || c | c c | c | c c}
      \hline
       & & & \multicolumn{3}{c |}{Test RMSE} & \multicolumn{3}{c}{Test log-likelihood} \\
      \bf{Dataset} & \bf{N} & \bf{D} & \bf{MC-Dropout} & \bf{BBVI} & \bf{Vadam} & \bf{MC-Dropout} & \bf{BBVI} & \bf{Vadam} \\
      \hline
      Boston   & 506 & 13 & 2.97 $\pm$ 0.19 & \bf{3.58 $\pm$ 0.21} & 3.93 $\pm$ 0.26 & -2.46 $\pm$ 0.06 & \bf{-2.73 $\pm$ 0.05} & -2.85 $\pm$ 0.07 \\
      Concrete & 1030 & 8 & 5.23 $\pm$ 0.12 & \bf{6.14 $\pm$ 0.13} & 6.85 $\pm$ 0.09 & -3.04 $\pm$ 0.02 & \bf{-3.24 $\pm$ 0.02} & -3.39 $\pm$ 0.02 \\
      Energy   & 768 & 8 & 1.66 $\pm$ 0.04 & 2.79 $\pm$ 0.06 & \bf{1.55 $\pm$ 0.08} & -1.99 $\pm$ 0.02 & -2.47 $\pm$ 0.02 & \bf{-2.15 $\pm$ 0.07} \\
      Kin8nm   & 8192 & 8 & 0.10 $\pm$ 0.00 & \bf{0.09 $\pm$ 0.00} & 0.10 $\pm$ 0.00 & 0.95 $\pm$ 0.01 & \bf{0.95 $\pm$ 0.01} & 0.76 $\pm$ 0.00 \\
      Naval    & 11934 & 16 & 0.01 $\pm$ 0.00 & \bf{0.00 $\pm$ 0.00} & \bf{0.00 $\pm$ 0.00} & 3.80 $\pm$ 0.01 & \bf{4.46 $\pm$ 0.03} & \bf{4.72 $\pm$ 0.22} \\
      Power    & 9568 & 4 & 4.02 $\pm$ 0.04 & 4.31 $\pm$ 0.03 & \bf{4.28 $\pm$ 0.03} & -2.80 $\pm$ 0.01 & \bf{-2.88 $\pm$ 0.01} & \bf{-2.88 $\pm$ 0.01} \\
      Wine     & 1599 & 11 & 0.62 $\pm$ 0.01 & \bf{0.65 $\pm$ 0.01} & 0.66 $\pm$ 0.01 & -0.93 $\pm$ 0.01 & \bf{-1.00 $\pm$ 0.01} & -1.01 $\pm$ 0.01 \\
      Yacht    & 308 & 6 & 1.11 $\pm$ 0.09 & 2.05 $\pm$ 0.06 & \bf{1.32 $\pm$ 0.10} & -1.55 $\pm$ 0.03 & -2.41 $\pm$ 0.02 & \bf{-1.70 $\pm$ 0.03} \\
      \hline
   \end{tabular}
\end{table*}

\begin{figure*}[t]
  \centering
  \subfigure{
     \includegraphics[width=0.19\linewidth]{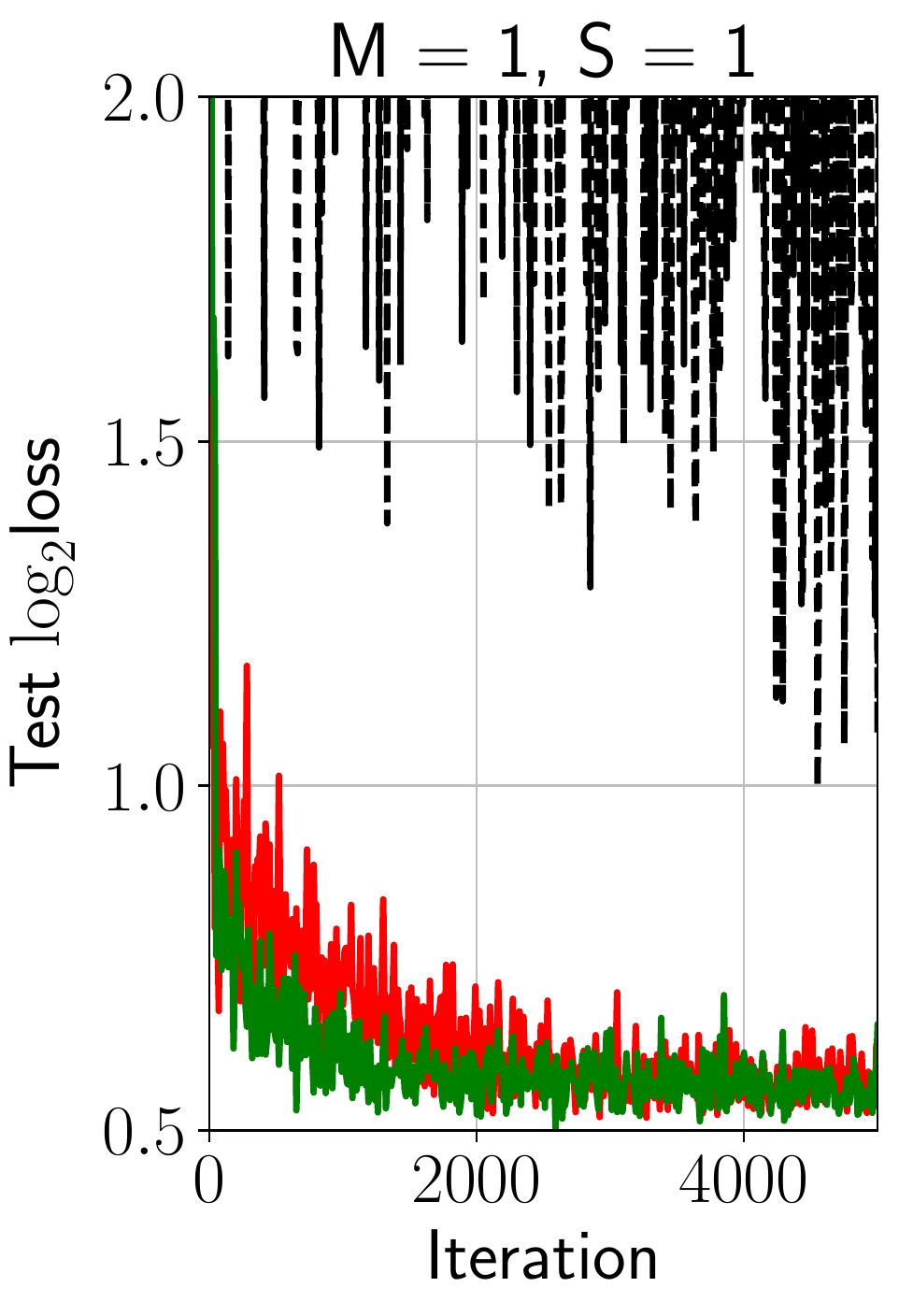}\includegraphics[width=0.19\linewidth]{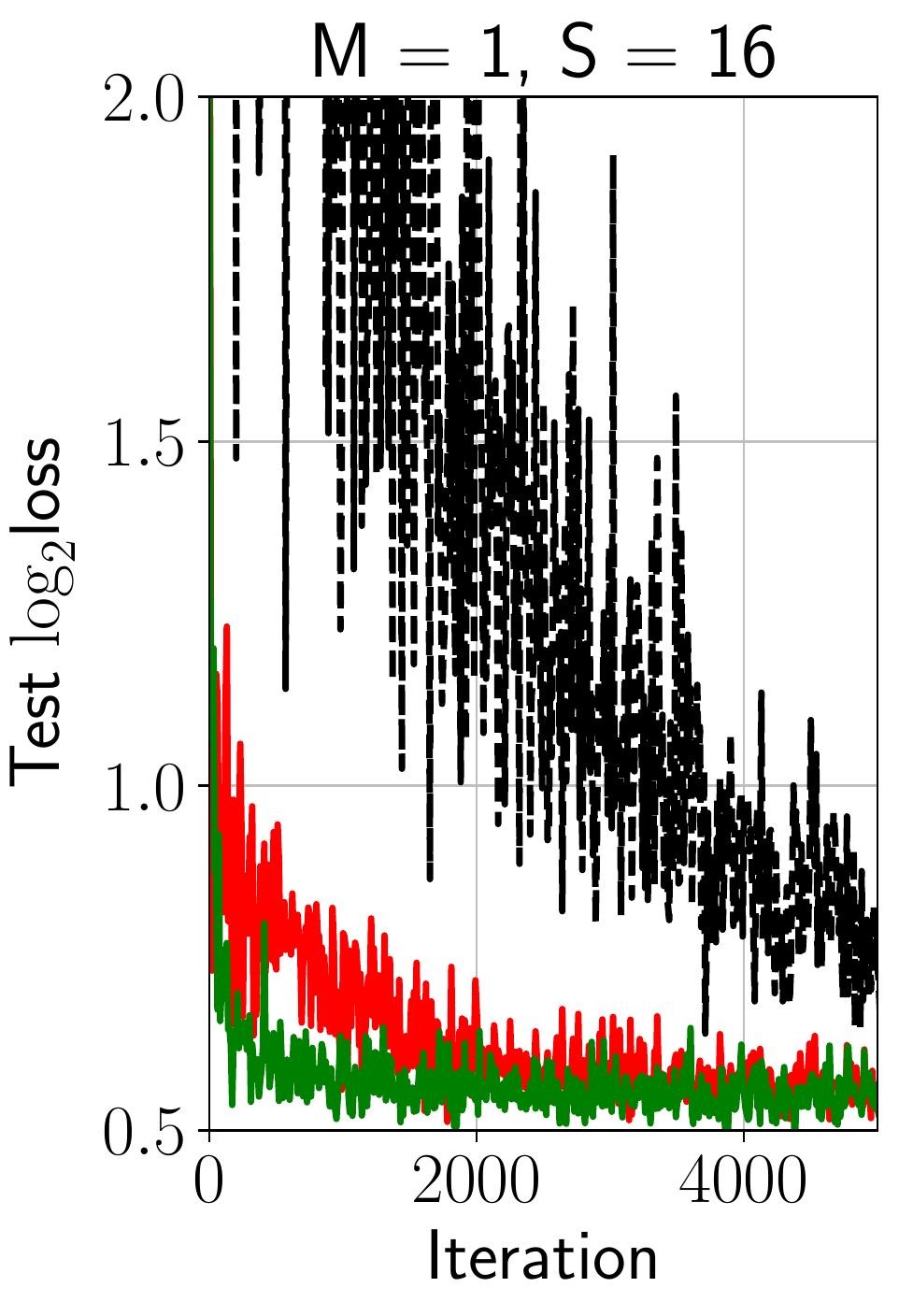}
  \includegraphics[width=0.19\linewidth]{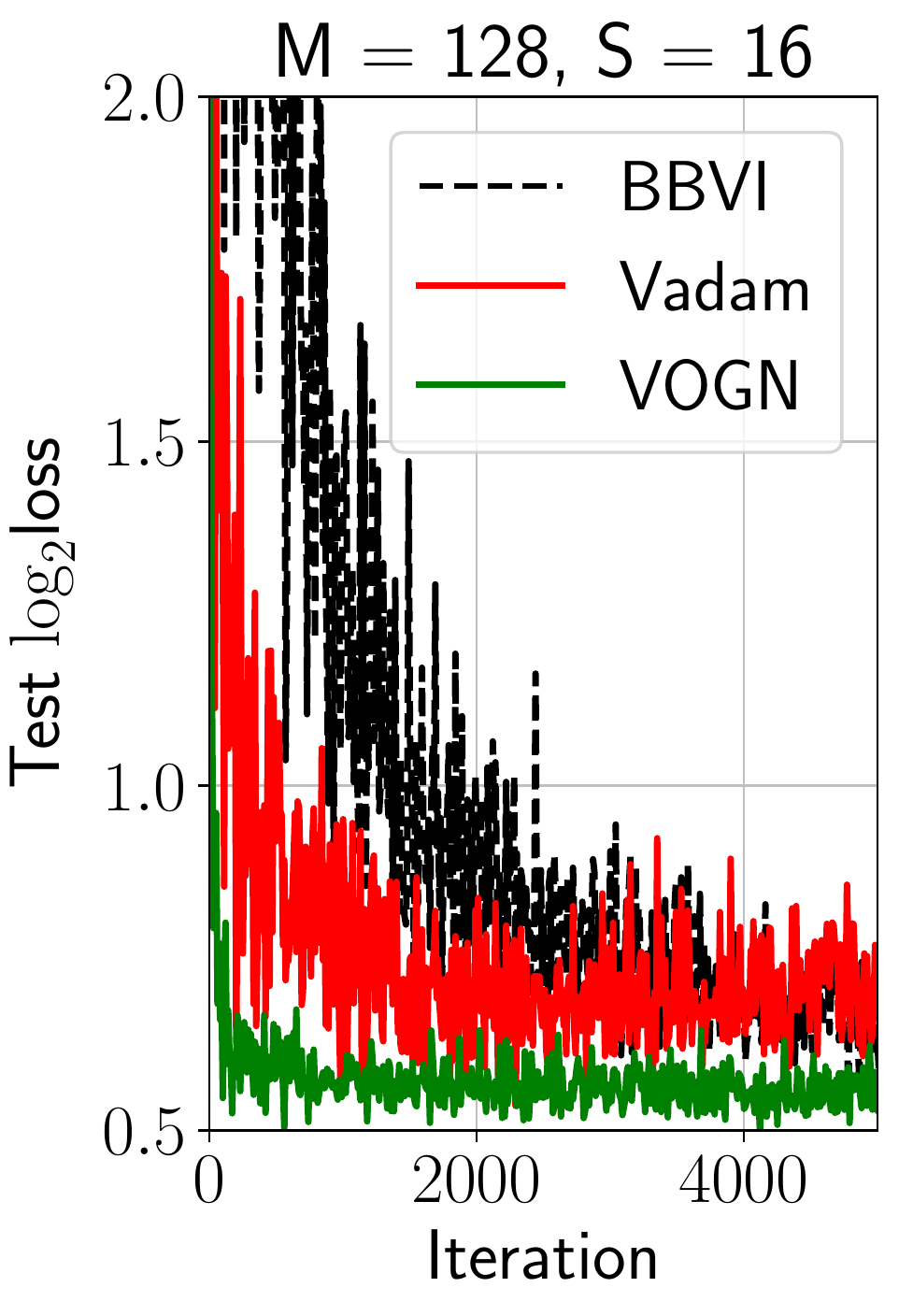}} 
  \hspace{0.1in}
  \subfigure{
     \includegraphics[width=0.36\linewidth]{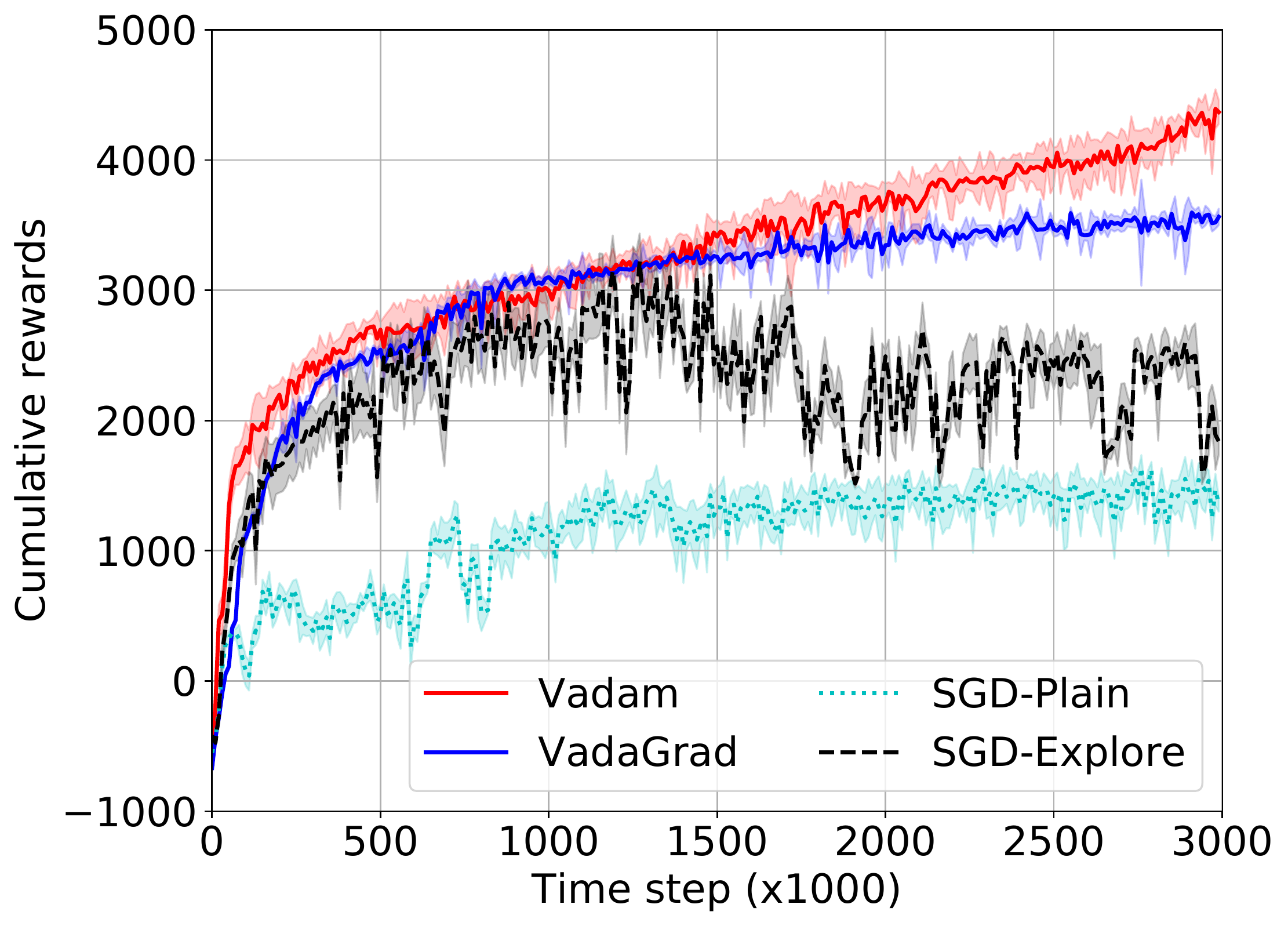}
  }
  \caption{The first 3 figures in the left show results on the Australian-Scale dataset using a neural network with a hidden layer of 64 units for different minibatch sizes $M$ and number of MC samples $S$. We see that VOGN converges the fastest, and Vadam too performs well for $M=1$. The rightmost figure shows results for exploration in deep RL where Vadam and VadaGrad outperform SGD-based methods.}
  \label{figure:results}
\end{figure*}

\subsection{Uncertainty Estimation in Neural Network}
\label{sec:bnn}
We show results on the standard UCI benchmark. We repeat the experimental setup used in~\citet{yarin16dropout}. Following their work, we use a neural network with one hidden layer, 50 hidden units, and ReLU activation functions. 
We use the 20 splits of the data provided by~\citet{yarin16dropout} for training and testing.
We use Bayesian optimization to select the prior precision $\lambda$ and noise precision of the Gaussian likelihood. 
Further details of the experiments are given in Appendix \ref{app:detailUCI}. 

We compare Vadam to MC-Dropout~\citep{yarin16dropout} using the results reported in~\citet{yarin16dropout}. We also compare to an SG method using the reparameterization trick and the Adam optimizer (referred to as `BBVI'). For a fair comparison, the Adam optimizer is run with the same learning rates as Vadam, although these can be tuned further to get better performance.

Table~\ref{table:uci} shows the performance in terms of the test RMSE and the test log-likelihood.
The better method out of BBVI and Vadam is shown in boldface found using a paired t-test with $p\textrm{-value}>0.01$. Both methods perform comparably, which supports our conclusion, however, MC-Dropout outperforms both the methods. We also find that VOGN shows similar results to Vadam and BBVI (we omit the results due to lack of space). The convergence plots for the final runs is given in Appendix~\ref{app:detailUCI}.

For many tasks, we find that VOGN and Vadam converge much faster than BBVI.
An example is shown in Figure \ref{figure:results} (see the first 3 figures in the left; details are in Appendix~\ref{app:detailUCI}). We have observed similar trends on other datasets.

\subsection{Exploration in Deep Reinforcement Learning}
\label{section:vo_rl}
A good exploration strategy is crucial in reinforcement learning (RL) since the data is sequentially collected. We show that weight-perturbation in Vadam improves exploration in RL. Due to space constraints, we only provide a brief summary of our results, and give details in Appendix~\ref{app:rl}.

We consider the deep deterministic policy gradient (DDPG) method for the Half-Cheetah task using a two-layer neural networks with $400$ and $300$ ReLU hidden units~\citep{DBLP:journals/corr/LillicrapHPHETS15}.
We compare Vadam and VadaGrad to two SGD methods, one of which does exploration (referred to as `SGD-Explore'), and the other does not (referred to as `SGD-plain'). The rightmost plot in Figure~\ref{figure:results} shows the cumulative rewards (higher is better) of each method against training iterations.
VadaGrad and Vadam clearly learn faster than both SGD-Plain and SGD-Explore.
We also compare the performances against Adam variants of SGD-Plain and SGD-Explore. Their results, given in the Appendix~\ref{app:rl}, show that Vadam and VadaGrad still learn faster, but only in the beginning and Adam based methods can catch up quickly.
This suggests that the exploration strategy has a high impact on the early learning performance in the Half-Cheetah task, and the effect of good exploration decreases over time as the agent collect more informative training samples.

\section{Discussion}
\label{sec:discussion}
In this paper, we present new VI algorithms which are as simple to implement and execute as algorithms for MLE. We obtain them by using a series of approximations and a natural momentum method for a natural-gradient VI method. The resulting algorithms can be implemented within Adam with minimal changes. Our empirical findings confirm that our proposed algorithms obtain comparable uncertainty estimates to existing VI methods, but require less computational and implementation effort\footnote{We made many new changes in this camera-ready version of the paper. A list of the changes is given in Appendix \ref{app:changes}.}.

An interesting direction we hope to pursue in the future is to generalize our natural-gradient approach to other types of approximation, e.g., exponetial-family distributions and their mixtures.
We would also like to further explore the application to areas such as RL and stochastic optimization.

\section*{Acknowledgements}
We thank the anonymous reviewers for their feedback.
We greatly appreciate many insightful discussions with Aaron Mishkin (UBC) and Frederik Kunstner (EPFL), and also thank them for their help on carrying out experiments and reviewing the manuscript.
We would also like to thank Roger Grosse and David Duvenaud from the University of Toronto for useful discussions.
We would like to thank Zuozhu Liu (SUTD, Singapore) for his help with the experiment on deep RL and logistic regression.
Finally, we are thankful for the RAIDEN computing system at the RIKEN Center for AI Project, which we extensively used for our experiments.

{\small
\bibliography{paper}
\bibliographystyle{icml2018}
}

\appendix
\onecolumn

\section{Changes in the Camera-Ready Version Compared to the Submitted Version}
\label{app:changes}
\begin{itemize}
   \item Taking reviewer's suggestions into account, we changed the title of our paper. The title of our submitted version was ``Vadam: Fast and Scalable Variational Inference by Perturbing Adam".
\item In the submitted version, we motivated our approach based on its ease of implementation. In the new version, we changed the motivation to make VI as easy to implement and execute as MLE.
\item In the new version, we have added a separate section on related work.
\item We improved the discussion of our approximation methods, and added an error analysis.
\item Overall conclusions of our paper have also slightly changed in the new version. The new conclusions suggest that there is a trade-off between the ease-of-implementation and quality of uncertainty approximation.
\item As per reviewers suggestions, we also made major improvements in our experiment results.
   \begin{itemize}
   \item We added test log-likelihood in the BNN results. We changed the hyperparameter selection from grid search to Bayesian optimization. We removed two methods from the table, namely PBP and VIG, since they use different splits compared to our setting. We improved the performance of BBVI by using better initialization and learning rates. We corrected a scaling problem with our Vadam method. The new method gives a slightly worse performance than the results present in
   the submitted version.
   \item We added a logistic regression experiment where we evaluate the quality of uncertainty estimates.
   \item We added the details of the RL experiments which we forgot to add in the submitted version. We also added a comparison to Adam-based methods in the appendix for the RL experiment.
   \item We removed an unclear result about reducing overfitting. 
   \end{itemize}
\item We added an additional result comparing VOGN with Vadam and BBVI on Bayesian neural network. 
\end{itemize}

\section{Review of Natural-Gradient Variational Inference}
\label{app:review}
\citet{khan2017conjugate} propose a natural-gradient method for variational inference. In this section, we briefly discuss this method.

Denote the variational objective by $\mathcal{L}(\veta)$ for the variational distribution $q_\eta(\vtheta)$ which takes an exponential-family form with natural-parameter $\veta$. The objective is given as follows:
\begin{align}
\mathcal{L}(\veta) := \sum_{i=1}^N \myexpect_q \sqr{\log p(\data_i|\vtheta)} + \myexpect_{q}\sqr{\log \frac{p(\vtheta)}{ q_\eta(\vtheta)}}. \label{eq:elbo_app}
\end{align}

We assume that the exponential-family is in minimal representation, which ensures that there is a one-to-one mapping between the natural parameter $\veta$ and the expectation parameter, denoted by $\vm$. Therefore, it is possible to express $\mathcal{L}(\veta)$ in terms of $\vm$. We denote this new objective by $\mathcal{L}_*(\vm) := \mathcal{L}(\veta)$. We can also reparameterize $q_\eta$ in terms of $\vm$ and denote it by $q_m$.

Natural-gradient methods exploit the Riemannian geometry of $q(\vtheta)$ by scaling the gradient by the inverse of the Fisher information matrix. The method of \citet{khan2017conjugate} simplifies the update by avoiding a direct computation of the FIM. This is made possible due to a relationship between the natural parameter $\veta$ and the expectation parameter $\vm$ of an exponential-family distribution. The relationship dictates that the natural gradient with respect to $\veta$ is equal to the gradient with respect to $\vm$. This is stated below where FIM is denoted by $\vF(\veta)$, 
\begin{align}
   \vF(\veta)^{-1} \nabla_\eta \mathcal{L}(\veta) = \nabla_m \mathcal{L}_*(\vm),
   \label{eq:natgradasmeangrad}
\end{align}
This relationship has been discussed in the earlier work of \citet{hensman2012fast} and can also be found in \citet{amari2016information}.

The method of \citet{khan2017conjugate} exploits this result within a mirror descent framework. They propose to use a mirror-descent update in the expectation- parameter space which is equivalent to the natural-gradient update in the natural-parameter space. Therefore, the natural-gradient update can be performed by using the gradient with respect to the expectation parameter. We give a formal statement below.
\begin{thm}
Consider the following mirror-descent step:
\begin{align}
   \vm_{t+1} = \arg\min_{\boldsymbol{m}}\,\, \left\langle \vm, -\nabla_m{\mathcal{L}_*(\vm_t)} \right\rangle + \frac{1}{\beta_t} \dkls{}{q_m(\vtheta)}{q_{m_t}(\vtheta)}, \label{eq:md1}
\end{align}
where $\dkls{}{\cdot}{\cdot}$ is the Kullback-Leibler divergence and $\beta_t$ is the learning rate in iteration $t$. Each step of this mirror descent update is equivalent to the following natural-gradient descent in the natural-parameter space:
\begin{align}
 \veta_{t+1} = \veta_t + \beta_t \vF(\veta_t)^{-1} \nabla_\eta \mathcal{L}(\veta_t)
\end{align}
\end{thm}
A formal proof of this statement can be found in \citet{raskutti2015information}.

Using \eqref{eq:natgradasmeangrad}, the natural-gradient update above can be simply written as the following:
\begin{align}
 \veta_{t+1} = \veta_t + \beta_t \nabla_m \mathcal{L}_*(\vm_t)
 \label{eq:md2}
\end{align}
which involves computing the gradient with respect to $\vm$ but taking a step in the natural-parameter space.

As we show in the next section, the above relationship enables us to derive a simple natural-gradient update because, for a Gaussian distribution, the gradient with respect to $\vm$ leads to a simple update.

\section{Derivation of Natural-Gradient Updates for Gaussian Mean-Field Variational Inference}
\label{app:mirror}
In this section, we derive the natural-gradient update for the Gaussian approximation $q_\eta(\vtheta) := \gauss(\vtheta|\vmu,\vSigma)$ with mean $\vmu$ and covariance matrix $\vSigma$. In the end, we will make the mean-field approximation: $\vSigma = \diag(\vsigma^2)$ to get the final update.

We start by defining the natural and expectation parameters of a Gaussian: 
\begin{align}
\veta^{(1)} := \vSigma^{-1}\vmu, \quad &\veta^{(2)} = -\half \vSigma^{-1} \\
                                       \vm^{(1)} := \myexpect_{q}(\vtheta) = \vmu,  \quad &\vM^{(2)} = \myexpect_{q}(\vtheta\vtheta^\top) = \vmu \vmu^\top + \vSigma.
\end{align}
Now we will express the gradient with respect to these expectation parameters in terms of the gradients with respect to $\vmu$ and $\vSigma$ using the chain rule (see Appendix B.1~in \citet{khan2017conjugate} for a derivation):
\begin{align}
\nabla_{m^{(1)}} \mathcal{L}_* &= \nabla_{\mu} \mathcal{L} - 2\left[\nabla_{\Sigma} \mathcal{L} \right]\vmu, \\
\nabla_{M^{(2)}} \mathcal{L}_* &= \nabla_{\Sigma} \mathcal{L}.
\end{align}
Next, using the definition of natural parameters, we can rewrite~\eqref{eq:md2} in terms of $\vmu$ and $\vSigma$ (here $\nabla_x \mathcal{L}_t$ implies that it is gradient of the variational objective with respect to a variable $\vx$ at $\vx=\vx_t$):
\begin{align}
\vSigma_{t+1}^{-1} &= \vSigma_t^{-1} - 2\beta_t \left[ \nabla_{\Sigma} \mathcal{L}_t \right] \\
\vmu_{t+1} &= \vSigma_{t+1} \left[ \vSigma_t^{-1}\vmu_t + \beta_t \left( \nabla_{\mu} \mathcal{L}_t - 2\left[\nabla_{\Sigma} \mathcal{L}_t \right] \vmu_t \right) \right] \\
&= \vSigma_{t+1} \left[ \vSigma_t^{-1}\vmu_t + \beta_t \nabla_{\mu} \mathcal{L}_t - 2\beta_t\left[\nabla_{\Sigma} \mathcal{L}_t \right] \vmu_t \right] \\
&= \vSigma_{t+1} \left[ \left( \vSigma_t^{-1} - 2\beta_t\left[\nabla_{\Sigma} \mathcal{L}_t \right] \right) \vmu_t + \beta_t \nabla_{\mu} \mathcal{L}_t \right] \\
&= \vSigma_{t+1} \left[ \vSigma^{-1}_{t+1} \vmu_t + \beta_t \nabla_{\mu} \mathcal{L}_t \right] \\
&= \vmu_t + \beta_t\vSigma_{t+1} \left[ \nabla_{\mu} \mathcal{L}_t \right].
\end{align}

In summary, the natural-gradient update is
\begin{align}
\vSigma_{t+1}^{-1} &= \vSigma_t^{-1} - 2\beta_t \left[ \nabla_{\Sigma} \mathcal{L}_t \right] , \\
\vmu_{t+1} &= \vmu_t + \beta_t\vSigma_{t+1} \left[ \nabla_{\mu} \mathcal{L}_t \right]. \label{eq:Van__0_app} 
\end{align}
By considering a Gaussian mean-field VI with a diagonal covariance: $\vSigma = \diag(\vsigma^2)$, we obtain
\begin{align}
\vsigma_{t+1}^{-2} &= \vsigma_t^{-2} - 2\beta_t \left[ \nabla_{\sigma^2} \mathcal{L}_t \right], \\
\vmu_{t+1} &= \vmu_t + \beta_t\vsigma^2_{t+1} \circ \left[ \nabla_{\mu} \mathcal{L}_t \right]. 
\end{align}
In update \eqref{eq:Van__0}, we use stochastic gradients instead of exact gradients.

Note that there is an explicit constraint in the above update, i.e., the precision $\vsigma^{-2}$ needs to be positive at every step. The learning rate can be adapted to make sure that the constraint is always satisfied. We discuss this method in Appendix \ref{app:hess_reparam}. Another option is to make an approximation, such as a Gauss-Newton approximation, to make sure that this constraint is always satisfied. We make this assumption in one of our methods called
Variational Online Gauss-Newton Method.

\section{Derivation of the Variational Online-Newton method (VON)}
\label{app:von}
In this section, we derive the variational online-Newton (VON) method proposed in Section~\ref{sec:vprop}. We will modify the NGVI update in~\eqref{eq:Van__0_app}. 

The variational lower-bound in~\eqref{eq:elbo_app} can be re-expressed as
\begin{align}
\mathcal{L}(\vmu,\vSigma) := \myexpect_q \sqr{ - N f(\vtheta) + \log p(\vtheta) - \log q(\vtheta) },
\end{align}
where $f(\vtheta) = -\frac{1}{N} \sum_{i=1}^N \log p(\mathcal{D}_i|\vtheta)$.
To derive VON, we use the Bonnet's and Price's theorems \cite{Opper:09,rezende2014stochastic} to express the gradients of the expectation of $f(\vtheta)$ with respect to $\vmu$ and $\vSigma$ in terms of the gradient and Hessian of $f(\vtheta)$, i.e.,  
\begin{align}
\nabla_{\mu} \myexpect_q \sqr{f(\vtheta)} &= \myexpect_q \sqr{ \nabla_\theta f(\vtheta)} := \myexpect_q \sqr{ \vg(\vtheta)} 
, \label{eq:bonnet}\\
\nabla_\Sigma \myexpect_q \sqr{f(\vtheta)} &= \half  \myexpect_q \sqr{ \nabla^2_{\theta\theta} f(\vtheta) } := \half \myexpect_q \sqr{ \vH(\vtheta)} ,
\end{align}
where ${\vg}(\vtheta) := {\nabla}_{\theta} f(\vtheta)$ and ${\vH}(\vtheta) := {\nabla}_{\theta\theta}^2 f(\vtheta)$ denote the gradient and Hessian of $f(\vtheta)$, respectively.
Using these, we can rewrite the gradients of $\mathcal{L}$ required in the NGVI update in~\eqref{eq:Van__0_app} as
\begin{align}
\nabla_\mu \mathcal{L} 
&= \nabla_\mu \myexpect_{q} \sqr{  - N f(\vtheta) + \log p(\vtheta) - \log q(\vtheta)} \\
&= - \left( \myexpect_{q}\sqr{ N\nabla_\theta f(\vtheta)} + \lambda\vmu \right)  \\
&= - \left( \myexpect_{q}\sqr{ N\vg(\vtheta)} + \lambda\vmu \right) ,  \label{eq:mfgradmu}\\
\nabla_{\Sigma} \mathcal{L} 
&= \half \myexpect_{q} \sqr{ - N \nabla^2_{\theta\theta} f(\vtheta) } - \half \lambda\vI + \half\vSigma^{-1} \\
&= \half \myexpect_{q} \sqr{ - N  \vH (\vtheta) } - \half \lambda\vI + \half\vSigma^{-1}, \label{eq:mfgradsig}
\end{align}
By substituting these into the NGVI update of \eqref{eq:Van__0_app} and then approximating the expectation by one Monte-Carlo sample $\vtheta_t \sim \gauss(\vtheta | \vmu_t,\vSigma_t)$, we get the following update:
\begin{align}
\vmu_{t+1} &= \vmu_{t} - \beta_t \,\, \vSigma_{t+1}  \sqr{  N\vg(\vtheta_t) + \lambda\vmu_t}, \label{eq:Van_mean_1}\\
\vSigma_{t+1}^{-1} &= (1-\beta_t) \vSigma_t^{-1} + \,\, \beta_t \,\, \sqr{ N \vH(\vtheta_t) + \lambda\vI}. \label{eq:Van__1}
\end{align}
By defining a matrix $\vS_t := (\vSigma_t^{-1} -\lambda\vI)/N$, we get the following:  
\begin{align}
\textrm{VON (Full-covariance): } &\vmu_{t+1} = \vmu_{t} - \beta_t \,\, \rnd{\vS_{t+1} + \lambda\vI/N}^{-1} \rnd{ \vg(\vtheta_t) + \lambda\vmu_t/N} , \nonumber \\
&\vS_{t+1} = (1-\beta_t) \vS_t +  \beta_t \,\,  \vH(\vtheta_t) , 
\label{eq:Von_prec_00}
\end{align}
where $\vtheta_t \sim \gauss(\vtheta|\vmu_t,\vSigma_t)$ with $\vSigma_t = [N(\vS_t + \lambda\vI/N)]^{-1}$.
We refer to this update as the Variational Online-Newton (VON) method because it resembles a regularized version of online Newton's method where the scaling matrix is estimated online using the Hessians.

For the mean-field variant, we can use a diagonal Hessian:
\begin{align}
\textrm{VON: } &\vmu_{t+1} = \vmu_{t} - \beta_t \,\, \rnd{ \vg(\vtheta_t) + \tlambda\vmu_t} / \rnd{\vs_{t+1} + \tlambda} , \nonumber \\
&\vs_{t+1} = (1-\beta_t) \vs_t +  \beta_t \,\,  \diag(\vH(\vtheta_t)) , 
\label{eq:Von_prec_00_diag}
\end{align}
where $\va/\vb$ denote the element-wise division operation between vectors $\va$ and $\vb$, and we have defined $\tlambda = \lambda/N$, and $\vtheta_t \sim \gauss(\vtheta|\vmu_t, \diag(\vsigma^2_t))$ with $\vsigma^2_t = 1/[N(\vs_t + \tlambda)]$.

By replacing $\vg$ and $\vH$ by their stochastic estimates, we obtain the VON update shown in~\eqref{eq:Von_prec_0} of the main text.

\subsection{Hessian Approximation Using the Reparameterization Trick}
\label{app:hess_reparam}

In this section we briefly discuss an alternative Hessian approximation approach for mean-field VI beside the generalized Gauss-Newton and gradient magnitude which are discussed in the main paper.
This approach is based on the reparameterization trick for the expectation of function over a Gaussian distribution.
By using the identity in~\eqref{eq:bonnet} for the mean-field case, we can derive a Hessian approximation using this:
\begin{align}
   \myexpect_q\sqr{\nabla_{\theta \theta}^2 f(\vtheta) } &= 2 \nabla_{\sigma^2} \myexpect_q[f(\vtheta)]  , \\
                                                         & = 2 \nabla_{\sigma^2} \myexpect_{\mathcal{N}(\epsilon|0,I)} \sqr{ f(\vmu + \vsigma \vepsilon) } \\
                                                         & = 2 \myexpect_{\mathcal{N}(\epsilon|0,I)}\sqr{  \nabla_{\sigma^2} f(\vmu + \vsigma \vepsilon) }  \\
                                                         & = \myexpect_{\mathcal{N}(\epsilon|0,I)}\sqr{  \nabla_{\theta} f(\vtheta)  \vepsilon/\vsigma }  \\
                                                         & \approx \hat{\vg}(\vtheta) \rnd{\vepsilon/\vsigma } , \label{eq:reparam2}
\end{align}
where $\vepsilon \sim \mathcal{N}(\vepsilon | 0, \vI)$ and $\vtheta = \vmu + \vsigma \vepsilon$.
By defining $\vs_t := \vsigma_t^{-2} -\lambda$, we can write the VON update using the reparameterization trick Hessian approximation as
\begin{align}
   \textrm{VON (Reparam): } 
 &\vmu_{t+1} = \vmu_{t} - \alpha_t\,\, \rnd{ \hat{\vg}(\vtheta_t) + \tlambda\vmu_t}/\rnd{\vs_{t+1} + \tlambda} \\
   &\vs_{t+1} = (1-\beta_t) \vs_t +  \beta_t \,\, \sqr{\hat{\vg}(\vtheta_t) ( \vepsilon_t / \vsigma_t) }, 
\end{align}
where $\vepsilon_t\sim\gauss(\vepsilon | 0,\vI)$ and $\vtheta_t = \vmu_t + \vepsilon_t/\sqrt{N(\vs_t + \tlambda)}$.

One major issue with this approximation is that it might have a high variance and $\vs_t$ may be negative. 
To make sure that $\vs_t > 0$ for all $t$, we can use a simple back-tracking method described below. 
Denote element $d$ of $\vs$ as $s_d$ and simplify notation by denoting $h_d$ to be the $d$'th element of $\hat{\vg}(\vtheta) \rnd{\vepsilon/\vsigma } $.
For $\vs$ to remain positive, we need $s_d + \beta_t h_d > 0, \forall d$. As $h_d$ can become negative, a too large value for $\beta_t$ will move $\vs$ out of the feasible set.
We thus have to find the largest value we can set $\beta_t$ to such that $\vs$ is still in the feasible set. Let $\mathcal{I}$ denote the indices $d$ for which $s_d + \beta_t h_d \leq 0$.
We can ensure that $\vs$ stays in the feasible set by setting
\begin{equation}
   \beta_t = \min \bigg\{\beta_0, \delta \min_{d \in \mathcal{I}} \frac{s_d}{|h_d|}\bigg\},
\end{equation}
where $\beta_0$ is the maximum learning rate and $0 < \delta < 1$ is a constant to keep $\vs$ strictly within the feasible set (away from the borders).
However, this back-tracking method may be computationally expensive and is not trivial to implement within the RMSProp and Adam optimizers.

\section{Derivation of Vadam}
\label{app:vadam_deriv}

\subsection{Adam as an Adaptive Heavy-Ball Method}
Consider the following update of Adam (in the pseudocode in the main text, we used $\gamma_2 = 1-\beta$):
\begin{equation}
	\begin{split}
      \textrm{Adam: }  \vu_{t+1} &= \gamma_1 \vu_t + (1-\gamma_1) \hat{\vg}(\vtheta_t) \\
      {\vs}_{t+1} &= (1-\beta) {\vs}_t +  \beta \,\, [\hat{\vg}(\vtheta_t)]^2   \\
      \hat{\vu}_{t+1} &= \vu_{t+1}/(1-\gamma_1^{t}) \\
      \hat{\vs}_{t+1} &= \vs_{t+1}/(1-(1-\beta)^{t}) \\
      \vtheta_{t+1} &= \vtheta_{t} - \alpha \,\, \hat{\vu}_{t+1}  /(\sqrt{\hat{\vs}_{t+1}} + \delta) ,
	\end{split}
    \label{eq:adam_app}
\end{equation}
This update can be expressed as the following \emph{adaptive} version of the Polyak's heavy ball\footnote{\citet{wilson2017marginal} do not add the constant $\delta$ in $\sqrt{\vs}_t$ but in Adam a small constant is added.} method as shown in \citet{wilson2017marginal},
\begin{align}
   \vtheta_{t+1} = \vtheta_t - \bar{\alpha}_t \sqr{ \frac{1}{\sqrt{\hat{\vs}_{t+1}}  + \delta} } \hat{\vg}(\vtheta_t) + \bar{\gamma}_t \sqr{ \frac{\sqrt{\hat{\vs}_t} + \delta }{\sqrt{\hat{\vs}_{t+1}} + \delta} } (\vtheta_t - \vtheta_{t-1}) ,
\label{eq:adam_scaled_momen}
\end{align}
where $\bar{\alpha}_t, \bar{\gamma}_t$ are appropriately defined in terms of $\gamma_1$ as shown below:
\begin{align}
\bar{\alpha}_t &:= \alpha \frac{1-\gamma_1}{1-\gamma_1^t}, \quad  \bar{\gamma}_t := \gamma_1\frac{1-\gamma_1^{t-1}}{1-\gamma_1^t}
\end{align}

We will now show that, by using natural gradients in the Polyak's heavy ball, we get an update that is similar to~\eqref{eq:adam_scaled_momen}.
This allows us to implement our approximated NGVI methods by using Adam.

\subsection{Natural Momentum for Natural Gradient VI}
We propose the following update:
\begin{align}
      \veta_{t+1} = \veta_t + \bar{\alpha}_t \widetilde{\nabla}_\eta \mathcal{L}_t + \bar{\gamma}_t (\veta_t -\veta_{t-1}),
	\label{eq:nat_momen_app}
	\end{align}
We can show that \eqref{eq:nat_momen_app} can be written as the following mirror descent extension of \eqref{eq:md1},
\begin{align}
   \vm_{t+1} = \arg\min_{\boldsymbol{m}}\,\, \left\langle \vm, -\nabla_m{\mathcal{L}_*(\vm_t)} \right\rangle + \frac{1}{\beta_t} \dkls{}{q_m(\vtheta)}{q_{m_t}(\vtheta)} - \frac{\alpha_t}{\beta_t} \dkls{}{q_m(\vtheta)}{q_{m_{t-1}}(\vtheta)}, 
   \label{eq:mirror1111}
\end{align}
where $\mathcal{L}_*$ refers to the variational lower-bound defined in Appendix \ref{app:review}, and $\alpha_t$ and $\beta_t$ are two learning rates defined in terms of $\bar{\alpha}_t$ and $\bar{\gamma}_t$.
The last term here is a \emph{natural momentum} term, which is very similar to the momentum term in the heavy-ball methods. For example, \eqref{eq:momen} can be written as the following optimization problem:
\begin{align}
\min_\theta \vtheta^T\nabla_\theta f_1(\vtheta_t) + \frac{1}{\beta_t} \|\vtheta-\vtheta_t\|^2 - \frac{\alpha_t}{\beta_t}  \|\vtheta-\vtheta_{t-1}\|^2.
\end{align}
In our natural-momentum method, the Euclidean distance is replaced by a KL divergence, which explains the name \emph{natural}-momentum.

Equivalence between \eqref{eq:mirror1111} to \eqref{eq:nat_momen_app} can be established by directly taking the derivative, setting it to zero, and simplifying:
\begin{align}
   &-\nabla_m \mathcal{L}_*(\vm_t) + \frac{1}{\beta_t} (\veta_{t+1} - \veta_t) -  \frac{\alpha_t}{\beta_t} (\veta_{t+1} - \veta_{t-1}) = 0 \\
   \Rightarrow \veta_{t+1} &= \frac{1}{1-\alpha_t} \veta_t + \frac{\beta_t}{1-\alpha_t} \nabla_{m} \mathcal{L}_*(\vm_t) - \frac{\alpha_t}{1-\alpha_t} \veta_{t-1},\\
                        &= \veta_t + \frac{\beta_t}{1-\alpha_t} \nabla_m \mathcal{L}_*(\vm_t) + \frac{\alpha_t}{1-\alpha_t} (\veta_t -\veta_{t-1}),
\label{eq:vadam_np}
\end{align}
where we use the fact that gradient of the KL divergence with respect to $\vm$ is qual to the difference between the natural parameters of the distributions \cite{raskutti2015information, khan2017conjugate}.
Noting that the gradient with respect to $\vm$ is the natural-gradient with respect to $\veta$. Therefore, defining $\bar{\alpha}_t := \beta_t/(1-\alpha_t)$ and $\bar{\gamma}_t := \alpha_t/(1-\alpha_t)$, we establish that mirror-descent is equivalent to the natural-momentum approach we proposed.

\subsection{NGVI with Natural Momentum for Gaussian Approximations}
We will now derive the update for a Gaussian approximation $q(\vtheta) := \gauss(\vtheta|\vmu,\vSigma)$.

Recalling that the mean parameters of a Gaussian $q(\vtheta) = \gauss(\vtheta|\vmu,\vSigma)$ are $\vm^{(1)} = \vmu$ and $\vM^{(2)} = \vSigma + \vmu\vmu ^T$.
Similarly to Appendix \ref{app:mirror}, By using the chain rule, we can express the gradient $\nabla_{m} \mathcal{L}_*$ in terms of $\vmu$ and $\vSigma$ as
\begin{align}
    \nabla_{m^{(1)}} \mathcal{L}_* &= \nabla_{\mu} \mathcal{L} - 2\left[\nabla_{\Sigma} \mathcal{L} \right]\vmu, \\
    \nabla_{M^{(2)}} \mathcal{L}_* &= \nabla_{\Sigma} \mathcal{L}.
\end{align}
Using the natural parameters of a Gaussian defined as $\veta^{(1)} = \vSigma^{-1}\vmu$ and $\veta^{(2)} = - \frac{1}{2}\vSigma^{-1}$, we can rewrite the update \eqref{eq:vadam_np} in terms of the update for $\vmu$ and $\vSigma$. 
First, the update for $\vSigma$ is obtained by plugging $\veta^{(2)} = -\frac{1}{2} \vSigma^{-1}$ into~\eqref{eq:vadam_np}:
\begin{align}
    \vSigma_{t+1}^{-1} &= \frac{1}{1-\alpha_t} \vSigma_t^{-1} - \frac{\alpha_t}{1-\alpha_t}  \vSigma_{t-1}^{-1} - \frac{2\beta_t}{1-\alpha_t} \left[ \nabla_{\Sigma} \mathcal{L}_t \right].  \label{eq:updateSig_00}
\end{align}
Now, for $\vmu$, we first plugging  $\veta^{(1)} = \vSigma^{-1}\vmu$ into~\eqref{eq:vadam_np} and then rearrange the update to express some of the terms as $\vSigma_{t+1}$:
\begin{align}
   \vSigma_{t+1}^{-1} \vmu_{t+1} &=  \frac{1}{1-\alpha_t} \vSigma_t^{-1}\vmu_t  -  \frac{\alpha_t}{1-\alpha_t} \vSigma_{t-1}^{-1}\vmu_{t-1}   + \frac{\beta_t}{1-\alpha_t} \left( \nabla_{\mu} \mathcal{L}_t - 2\left[\nabla_{\Sigma} \mathcal{L}_t \right] \vmu_t \right)  \\
   &=  \frac{1}{1-\alpha_t} \vSigma_t^{-1}\vmu_t  -  \frac{\alpha_t}{1-\alpha_t} \vSigma_{t-1}^{-1}\vmu_{t-1}   + \frac{\beta_t}{1-\alpha_t} \left( \nabla_{\mu} \mathcal{L}_t - 2\left[\nabla_{\Sigma} \mathcal{L}_t \right] \vmu_t \right) +  \frac{\alpha_t}{1-\alpha_t}\left( \vSigma_{t-1}^{-1}\vmu_t - \vSigma_{t-1}^{-1}\vmu_t \right)\\
    &= \left[  \frac{1}{1-\alpha_t} \vSigma_t^{-1} -  \frac{\alpha_t}{1-\alpha_t} \vSigma_{t-1}^{-1}  -  \frac{2\beta_t}{1-\alpha_t} \left[ \nabla_{\Sigma} \mathcal{L}_t \right] \right]\vmu_t + \frac{\beta_t}{1-\alpha_t} \nabla_{\mu} \mathcal{L}_t  + \frac{\alpha_t}{1-\alpha_t}  \vSigma_{t-1}^{-1} (\vmu_{t}-\vmu_{t-1})  \\
   \Rightarrow \vmu_{t+1}   &= \vmu_t + \frac{\beta_t}{1-\alpha_t}\vSigma_{t+1} \left[ \nabla_{\mu} \mathcal{L}_t \right] + \frac{\alpha_t}{1-\alpha_t} \vSigma_{t+1} \vSigma_{t-1}^{-1}(\vmu_{t}-\vmu_{t-1}), 
    \label{eq:updateMu_00}
\end{align}
where in the final step, we substitute the definition of $\vSigma_{t+1}^{-1}$ from \eqref{eq:updateSig_00}. 

To express these updates similar to VON, we make an approximation where we replace the instances of $\vSigma_{t-1}$ by $\vSigma_t$ in both \eqref{eq:updateSig_00} and \eqref{eq:updateMu_00}. With this approximation, we get the following:
\begin{align} 
   \vmu_{t+1} &= \vmu_t + \frac{\beta_t}{1-\alpha_t}\vSigma_{t+1} \left[ \nabla_{\mu} \mathcal{L}_t \right] + \frac{\alpha_t}{1-\alpha_t} \vSigma_{t+1} \vSigma_{t}^{-1}(\vmu_{t}-\vmu_{t-1})  \label{eq:approx11}, \\
\vSigma_{t+1}^{-1} &= \vSigma_t^{-1} - \frac{2\beta_t}{1-\alpha_t} \left[ \nabla_{\Sigma} \mathcal{L}_t \right].
\end{align}
We build upon this update to express it as VON update with momentum.

\subsection{Variational Online Newton with Natural Momentum}
\label{eq:von_momen}

Now, we derive VON with natural momentum.
To do so, we follow the same procedure used to derive the VON update in Section \ref{app:von}. That is, we first use Bonnet's and Price's theorem to express the gradients with respect to $\vmu$ and $\vSigma$ in terms of the expectations of gradients and Hessian of $f(\vtheta)$. Then, we substitute the expectation with a sample $\vtheta_t \sim \gauss( \vtheta | \vmu_t, \vSigma_t)$. Finally, we redefine the matrix $\vS_t := (\vSigma_t^{-1} -\lambda\vI)/N$. With this we get the following update which is a momentum version of VON:
\begin{align}
 \vmu_{t+1} &= \vmu_{t} - \frac{\beta_t}{1-\alpha_t} \,\, \rnd{\vS_{t+1} + \tlambda\vI}^{-1} \rnd{ \vg(\vtheta_t) + \tlambda\vmu_t}  + \frac{\alpha_t}{1-\alpha_t} (\vS_{t+1} + \tlambda\vI)^{-1} (\vS_t + \tlambda\vI) (\vmu_t - \vmu_{t-1}), \\
 \vS_{t+1} &= \rnd{1-\frac{\beta_t}{1-\alpha_t}} \vS_t +  \frac{\beta_t}{1-\alpha_t} \,\, \vH(\vtheta_t), 
\end{align}
where $\vtheta_t \sim \gauss(\vtheta|\vmu_t,\vSigma_t)$ with $\vSigma_t = [N(\vS_t + \tlambda\vI)]^{-1}$.

To get a momentum version of Vprop, we follow a similar method to Section \ref{sec:vprop}. 
That is, we first employ a mean-field approximation, and then replace the Hessian by the gradient-magnitude approximation.
Doing so gives us
\begin{align}
\vmu_{t+1} &= \vmu_{t} - \frac{\beta_t}{1-\alpha_t}  \,\, \sqr{ \frac{1}{{\vs_{t+1}}  + \tlambda} } \rnd{ {\vg}(\vtheta_t) + \tlambda\vmu_t}  + \frac{\alpha_t}{1-\alpha_t}  \sqr{ \frac{{\vs_t} + \tlambda}{{\vs_{t+1}} + \tlambda} } (\vmu_t - \vmu_{t-1}), \\
\vs_{t+1} &= \rnd{1-\frac{\beta_t}{1-\alpha_t}} \vs_t +  \frac{\beta_t}{1-\alpha_t} \,\,  \sqr{{\vg}(\vtheta_t)}^2,    
\end{align}
where $\vtheta_t \sim \gauss(\vtheta|\vmu_t,\vsigma_t^2)$ with $\vsigma_t^2 = 1/[N(\vs_t + \tlambda)]$.
Finally, we use an unbiased gradient estimate $\hat{\vg}(\vtheta)$, introduce the square-root for the scaling vector in the mean update, and define step-sizes $\bar{\alpha}_t := \beta_t/(1-\alpha_t)$ and $\bar{\gamma}_t := \alpha_t/(1-\alpha_t)$.
The result is a \emph{Vprop with momentum} update:
 \begin{align}
 \vmu_{t+1} &= \vmu_{t} - \bar{\alpha}_t \,\, \sqr{ \frac{1}{\sqrt{\vs_{t+1}}  + \tlambda} } \rnd{ \hat{\vg}(\vtheta_t) + \tlambda\vmu_t}  + \bar{\gamma}_t \sqr{ \frac{\sqrt{\vs_t} + \tlambda}{\sqrt{\vs_{t+1}} + \tlambda} } (\vmu_t - \vmu_{t-1}), \\
 \vs_{t+1} &= \rnd{1-\bar{\alpha}_t} \vs_t +  \bar{\alpha}_t \,\,  \sqr{\hat{\vg}(\vtheta_t)}^2,        
 \end{align}
 where $\vtheta_t \sim \gauss(\vtheta|\vmu_t,\vsigma_t^2)$ with $\vsigma_t^2 = 1/[N(\vs_t + \tlambda)]$.
 This is very similar to the update \eqref{eq:adam_scaled_momen} of Adam expressed in the momentum form. 
 By introducing the bias correction term for $\vm$ and $\vs$, we can implement this update by using Adam's update shown in Fig. \ref{fig:adamVsVadam}. 
 The final update of Vadam is shown below, where we highlight the differences from Adam in red.
 \begin{align}
    \textcolor{red}{\vtheta_t} & \textcolor{red}{\sim \gauss(\vtheta|\vmu_t,  1/[N(\vs_t + \tlambda)] ) }, \\
      \vu_{t+1} &= \gamma_1 \vu_t + (1-\gamma_1) \left( \hat{\vg}(\vtheta_t) + \textcolor{red}{\tlambda \vmu_t} \right) \\
      {\vs}_{t+1} &= (1-\beta) {\vs}_t +  \beta \,\, [\hat{\vg}(\vtheta_t)]^2   \\
      \hat{\vu}_{t+1} &= \vu_{t+1}/(1-\gamma_1^{t}) \\
      \hat{\vs}_{t+1} &= \vs_{t+1} /(1-(1-\beta)^{t}) \\
      \vmu_{t+1} &= \vmu_{t} - \alpha \,\, \hat{\vu}_{t+1}  /(\sqrt{\hat{\vs}_{t+1}} + \textcolor{red}{\tlambda} ).
    \label{eq:vadam_app}
\end{align}
Note that we do not use the same step-size $\bar{\alpha}_t$ for $\vs_t$ and $\vmu_t$, but rather choose the step-sizes according to the Adam update. In the pseudocode, we define $\gamma_2 = 1-\beta$.

\section{The VadaGrad Update}
\label{app:vadagrad}
By setting $\tau=0$ in \eqref{eq:mu123}, we get the following update:
\begin{align}
   &\vmu_{t+1} = \vmu_{t} - \alpha_t\,\, \sqr{\widehat{\nabla}_\theta F(\vtheta)/\vs_{t+1}} ,  \\
   &\vs_{t+1} = \vs_t +  \beta_t \,\,  \widehat{\nabla}_{\theta\theta}^2 F(\vtheta), 
\end{align}
where $\vtheta_t \sim \gauss(\vtheta|\vmu_t,\vsigma_t^2)$ with $\vsigma^2_t := 1/\vs_t$.
By replacing the Hessian by a GM approximation, and taking the square-root as in Vprop, we get the following update we call VadaGrad:
\begin{align}
   &\vmu_{t+1} = \vmu_{t} - \alpha_t\,\, \sqr{\widehat{\nabla}_\theta F(\vtheta)/\sqrt{\vs_{t+1}}} , \nonumber\\
   &\vs_{t+1} = \vs_t +  \beta_t \,\, \sqr{\widehat{\nabla}_\theta F(\vtheta) \circ \widehat{\nabla}_\theta F(\vtheta) }, \label{eq:VadaGrad} 
\end{align}
where $\vtheta_t \sim \gauss(\vtheta|\vmu_t,\vsigma_t^2)$ with $\vsigma^2_t := 1/\vs_t$.

\section{Proof of Theorem 1}
\label{app:proof1}
Let $\vg_i := \nabla_{\theta} f_i(\vtheta)$ denote the gradient for an individual data point,
$\vg_{\minibatch} := \frac{1}{M} \sum_{i\in\minibatch} \nabla_{\theta} f_i(\vtheta)$ denote the average gradient over a minibatch of size $M$ and $\vg = \frac{1}{N} \sum_{i=1}^N \nabla_{\theta} f_i(\vtheta)$ denote the average full-batch gradient.
Let $p(i)$ denote a uniform distribution over the data samples $\{1,2,...,N\}$ and $p(\minibatch)$ a uniform distribution over the ${N \choose M}$ possible minibatches of size $M$.
Let further $\vG$ denote the average GGN matrix,
\begin{equation}
\vG = \frac{1}{N} \sum_{i=1}^N \vg_i \vg_i^T = \myexpect_{p(i)} [\vg_i \vg_i^T].
\end{equation}
Using the following two results,
\begin{align}
\Cov_{p(i)} [\vg_i ] &= \myexpect_{p(i)} [\vg_i \vg_i^T] - \myexpect_{p(i)}[\vg_i]\myexpect_{p(i)}[\vg_i]^T = \vG - \vg\vg^T,\\
\Cov_{p(\minibatch)} [\vg_{\minibatch} ] &= \myexpect_{p(\minibatch)} [\vg_{\minibatch} \vg_{\minibatch}^T] - \myexpect_{p(\minibatch)}[\vg_{\minibatch}]\myexpect_{p(\minibatch)}[\vg_{\minibatch}]^T = \myexpect_{p(\minibatch)} [\vg_{\minibatch} \vg_{\minibatch}^T] - \vg\vg^T,
\end{align}
along with Theorem 2.2 of \citet{cochran77} which states that
\begin{equation}
   \Cov_{p(\minibatch)} [\vg_{\minibatch} ] = \frac{1-\frac{M}{N}}{M}\frac{N}{N-1}\Cov_{p(i)} [\vg_i ],
\end{equation}
we get the following:
\begin{equation}
\myexpect_{p(\minibatch)} [\vg_{\minibatch} \vg_{\minibatch}^T] = w \vG + (1-w)\vg\vg^T,
\end{equation}
where $w = \frac{1}{M}(N-M)/(N-1)$.

Denoting dimension $j$ of the full-batch gradient by $\vg_j(\vtheta)$, dimension $j$ of the average gradient over a minibatch by $\hat{\vg}_j(\vtheta;\minibatch)$ and dimension $j$ of the diagonal of the average GGN, we get the stated result.

\section{Proof to Show That Fixed-Point of Vprop Do Not Change with Square-root}
\label{app:fixed}

We now show that the fixed-points do not change when we take the square root of $\vs_{t+1}$.
Denote the variational distribution at iteration $t$ by $q_t := \gauss(\vtheta|\vmu_t,\vsigma_t^2)$. Assume no stochasticity, i.e., we compute the full-batch gradients and also can exactly compute the expectation with respect to $q$.

A fixed point $q_*(\vtheta) := \gauss(\vtheta|\vmu_*,\vsigma_*^2)$ of the variational objective satisfies the following:
\begin{align}
N\myexpect_{q_*}\sqr{\nabla_\theta f(\vtheta)} + \lambda \vmu_* = 0,  \quad
N\myexpect_{q_*}\sqr{\diag\rnd{\nabla_{\theta\theta}^2 f(\vtheta)} } + \lambda - \vsigma_*^2 = 0, 
\end{align}

If we replace the Hessian by the GM approximation, we get the following fixed-point:
\begin{align}
N\myexpect_{q_*}\sqr{\nabla_\theta f(\vtheta)} + \lambda \vmu_* = 0,  \quad
N\myexpect_{q_*}\sqr{\rnd{\nabla_\theta f(\vtheta)}^2 } + \lambda - \vsigma_*^2 = 0, 
\end{align}
This fixed-point does not depend on the fact whether we scale by using the square-root or not. However, the iterations do depend on it and the scaling is expected to affect the convergence and also the path that we take to approach the solution.

\section{Details for the Logistic Regression Experiments}
\label{app:logreg_detail}
\subsection{Toy Example}

We used the toy example given in \citet{Murphy:2012:MLP:2380985} (see Fig. 8.6~in the book). The data is generated from a mixture of two Gaussians (details are given in the book). We used the generating mechansim desribed in the book to generate $N=60$ examples. For all methods, a prior precision of $\lambda = 0.01$ and $1$ MC sample is used. 
The initial settings of all methods are $\alpha_0 = 0.1$ and $\beta_0 = 0.9$. For every iteration $t$, the learning rates are decayed as
\begin{align}
\alpha_t &= \frac{\alpha_0}{1 + t^{0.55}}, \quad\quad 
\beta_t = 1- \frac{1-\beta_0}{1 + t^{0.55}}.
\label{eq:lr_decay}
\end{align}
Vadam and VOGN are run for 83,333 epochs using a minibatch size of $M=10$ (corresponding to 500,000 iterations). For Vadam, $\gamma_1$ is set to $\beta_t$.
VOGN-1 is run for 8000 epochs with a minibatch size of $M=1$ (also corresponding to 500,000 iterations).

\subsection{Real-Data Experiments}
Datasets for logistic regression are available at {\footnotesize \url{https://www.csie.ntu.edu.tw/~cjlin/libsvmtools/datasets/binary.html}}. For the Breast Cancer dataset, we use the hyper-parameters found by \citet{khan2017conjugate}. For USPS, we used the procedure of \citet{khan2017conjugate} to find the hyperparameter. All details are given in Table \ref{data_stat}. For all datasets we use 20 random splits.

\begin{table}[h]
\center
\caption{Datasets for logistic regression. $N_{\text{Train}}$ is the number of training data. 
}
\begin{tabular}{llllll}
\hline
Dataset & $N$ & $D$ & $N_{\text{Train}}$   & Hyperparameters & $M$ \\
\hline
USPS3vs5 & 1,781 & 256 & 884 & $\lambda=25$ & 64 \\
Breast-cancer-scale & 683  &  10 & 341  &   $\lambda=1.0$ & 32 \\
\hline
\end{tabular}
\label{data_stat}
\end{table}

{\bf Performance comparison of MF-Exact, VOGN-1, Vadam:}
We used 20 random 50-50 splits of the USPS 3vs5 dataset. For all methods, a prior precision of $\lambda = 25$ is used. MF-Exact and Vadam are run for $10000$ epochs with a minibatch size of $M=64$. The learning rates for both methods are decayed according to \eqref{eq:lr_decay} with initial settings $\alpha_0 = 0.01$ and $\beta_0 = 0.99$.
For Vadam, $1$ MC sample is used. 
VOGN-1, on the other hand, is run for $200$ epochs with a minibatch size of $M=1$, using $1$ MC sample and learning rates $\alpha=0.0005$ and $\beta=0.9995$.

{\bf Minibatch experiment comparing VOGN-1 and Vadam:} We use the Breast-Cancer dataset with 20 random initializations.  For both VOGN-1 and Vadam, a prior precision of $\lambda = 1$ is used. For VOGN-1, the learning rates are set to $\alpha = 0.0005$ and $\beta = 0.9995$. It is run for $2000$ epochs using a minibatch size of $M=1$ and 1 MC sample. For Vadam, the learning rates are decayed according to \eqref{eq:lr_decay} with initial settings $\alpha_0 = 0.01$ and $\beta_0 = 0.99$. The method is run with 1 MC sample for various minibatch sizes $M\in\{1,8,16,32,64\}$.

\section{Details for the Bayesian Neural Network Experiment}
\label{app:detailUCI}
\subsection{UCI Regression Experiments}

\begin{figure}[t]
\centering
\subfigure{\includegraphics[width=0.4\linewidth]{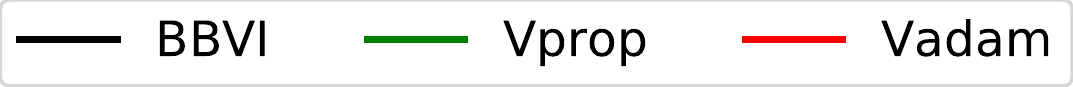}}
\hfill \\
\subfigure{\includegraphics[width=0.24\linewidth]{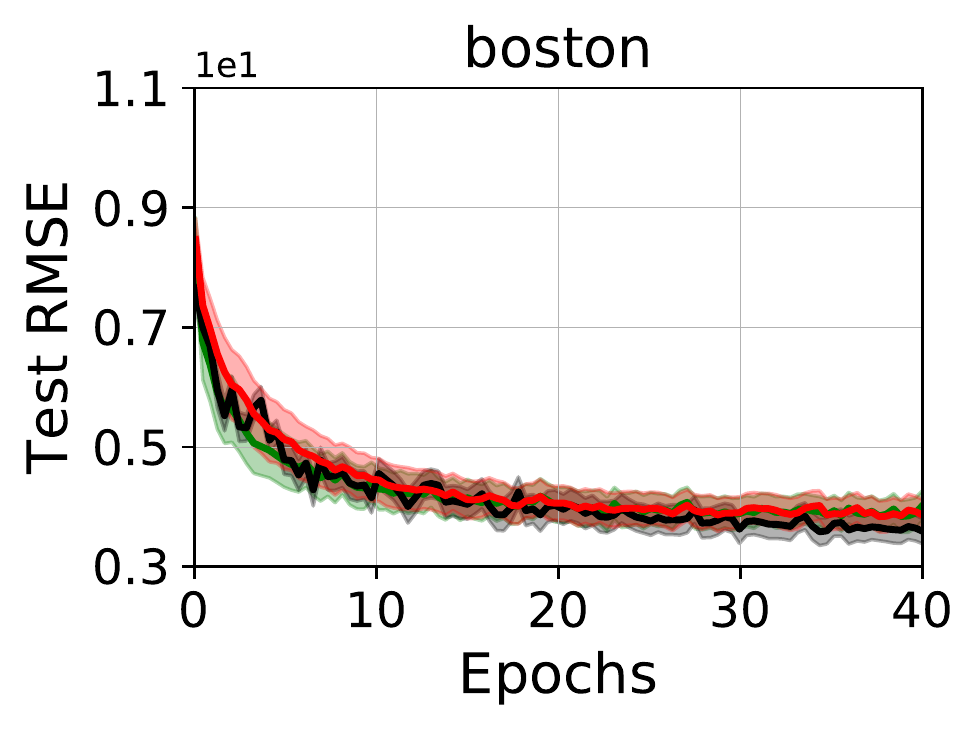}}
\subfigure{\includegraphics[width=0.24\linewidth]{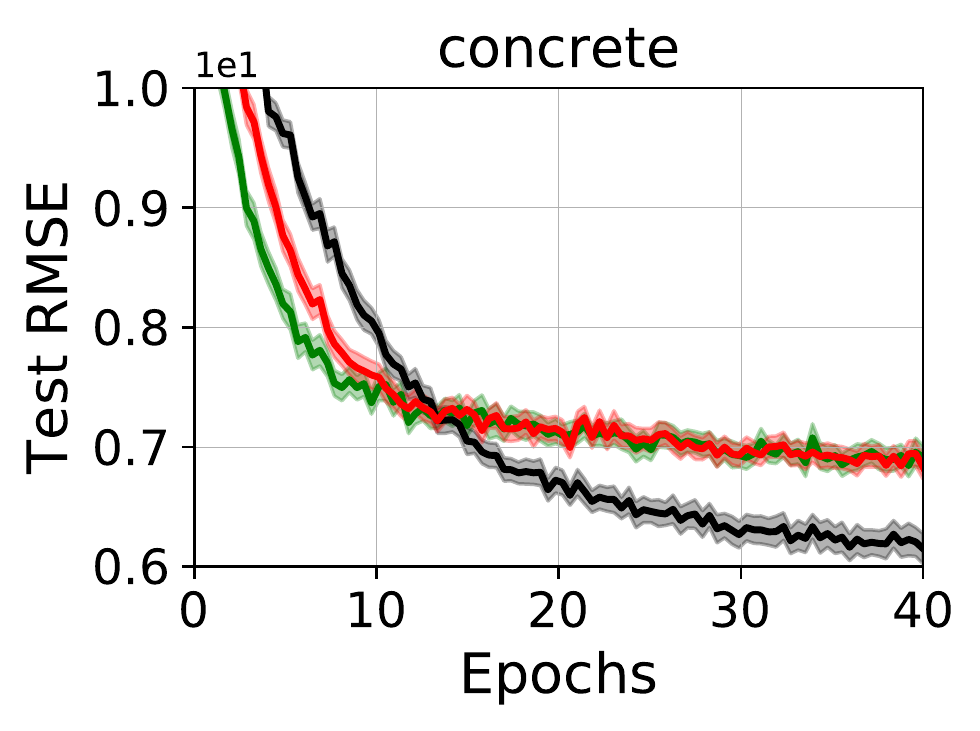}}
\subfigure{\includegraphics[width=0.24\linewidth]{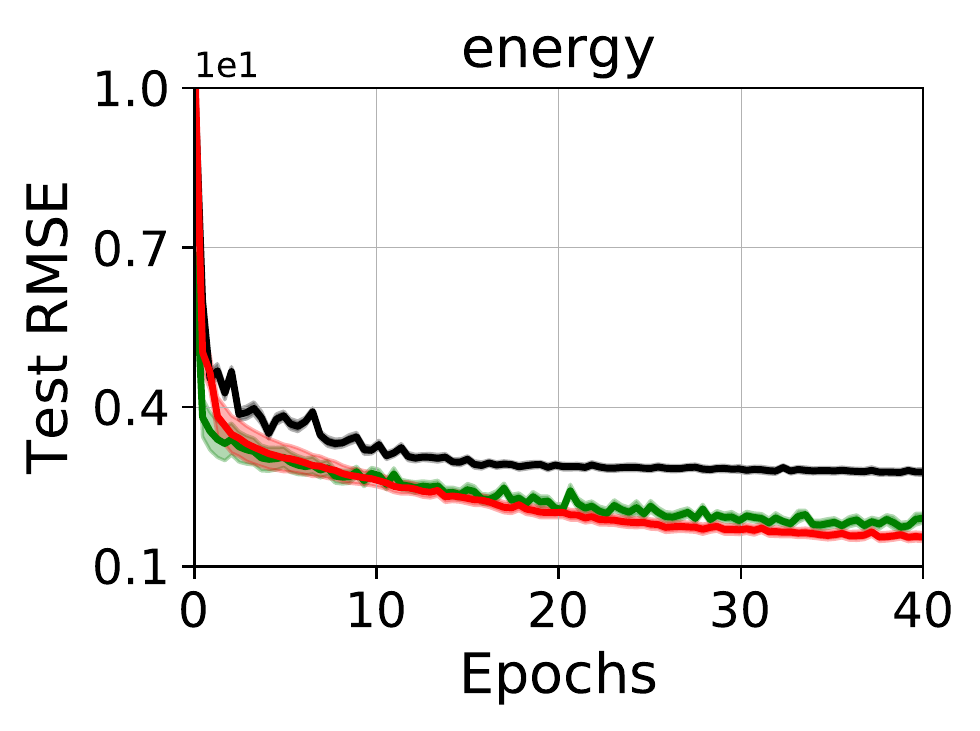}}
\subfigure{\includegraphics[width=0.24\linewidth]{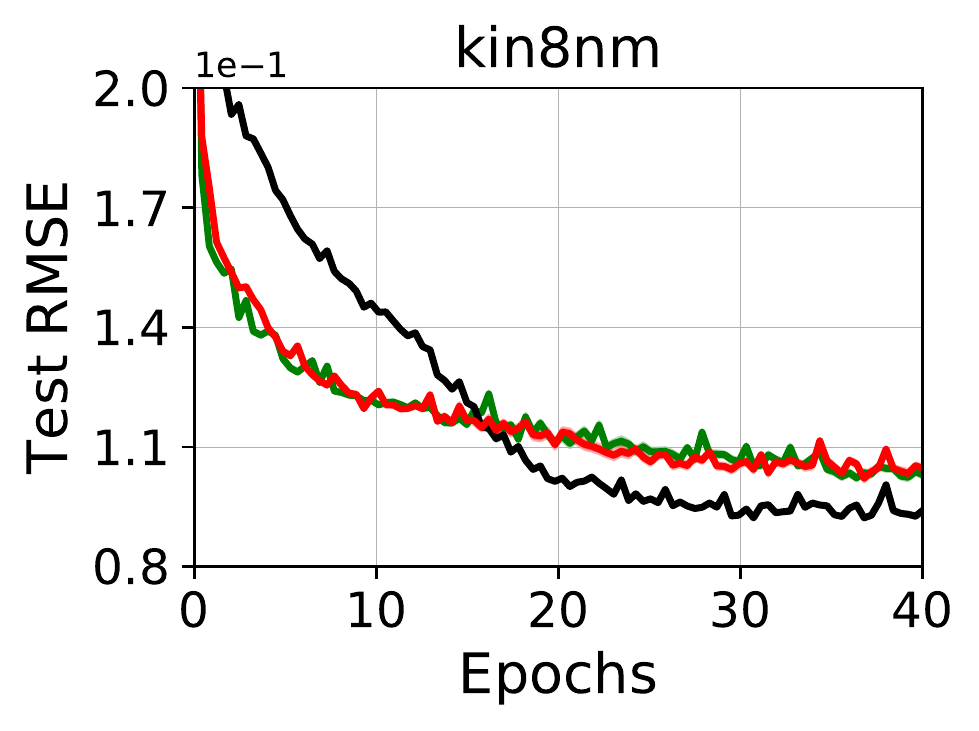}}
\hfill \\  
\centering
\subfigure{\includegraphics[width=0.24\linewidth]{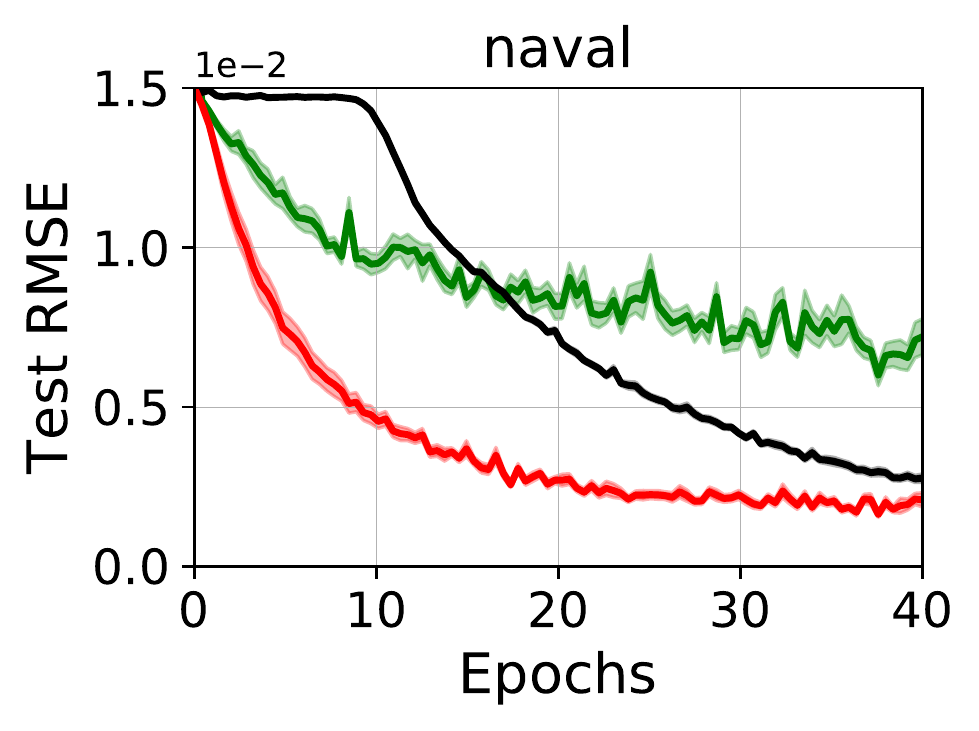}}
\subfigure{\includegraphics[width=0.24\linewidth]{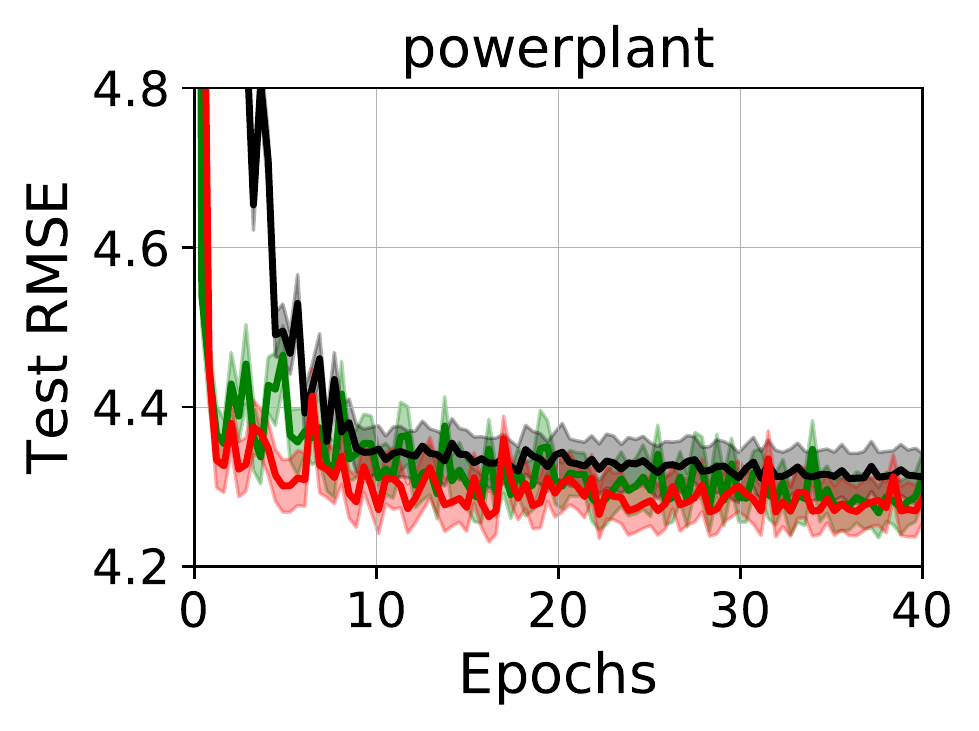}}
\subfigure{\includegraphics[width=0.24\linewidth]{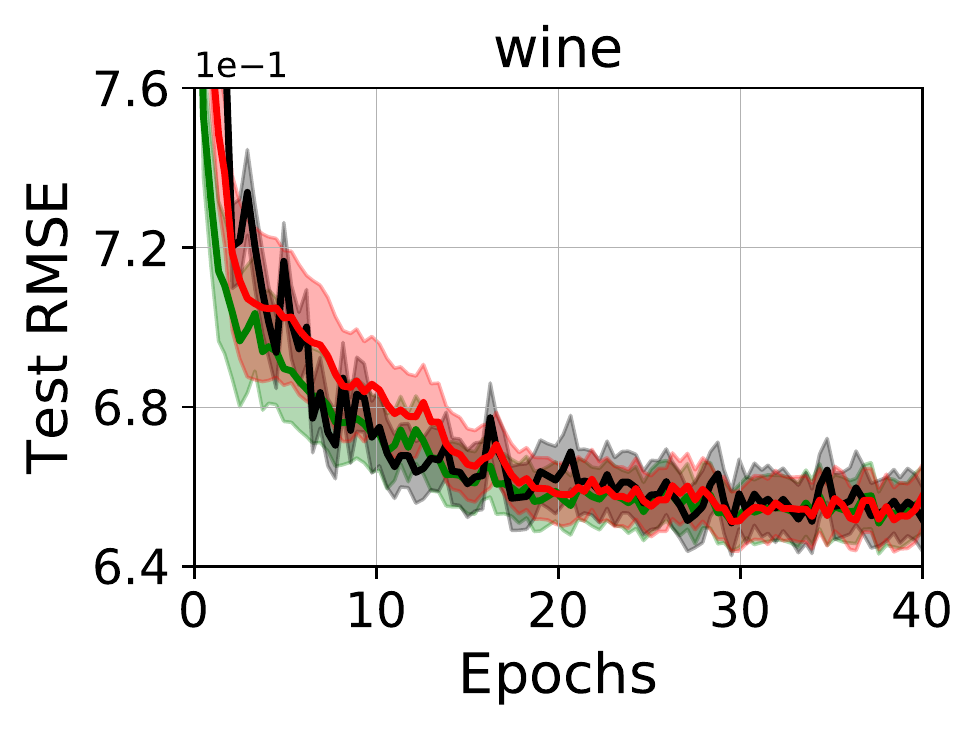}}
\subfigure{\includegraphics[width=0.24\linewidth]{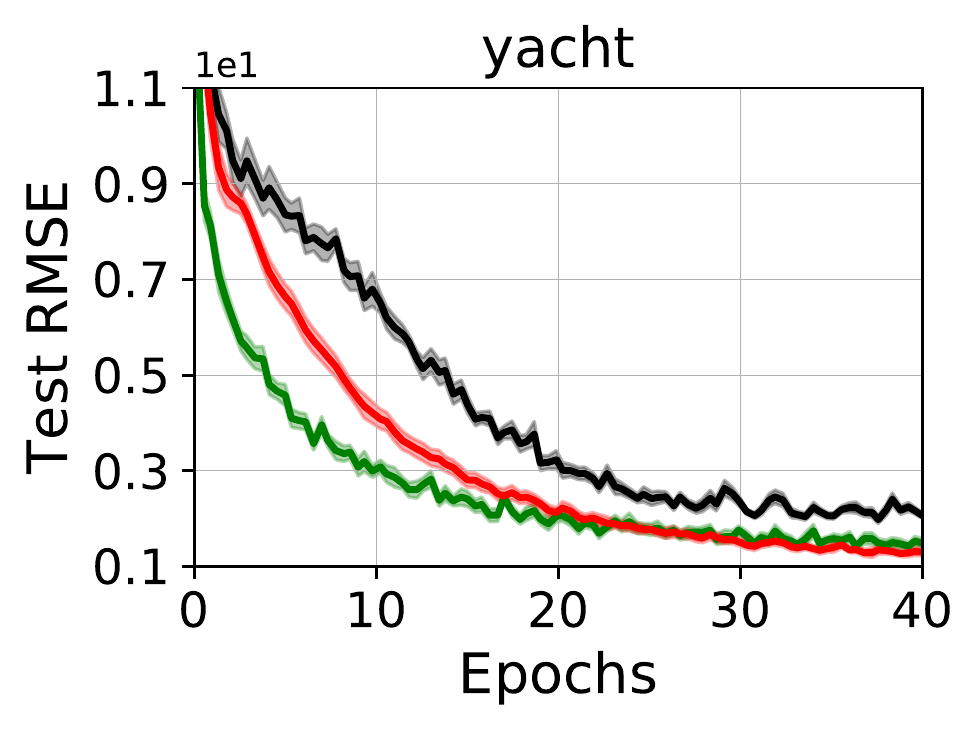}}
\caption{The mean plus-minus one standard error of the Test RMSE (using 100 Monte Carlo samples) on the test sets of UCI experiments. The mean and standard errors are computed over the 20 data splits.}
\label{figure:uci_regression_full}
\end{figure}

The 8 datasets together with their sizes $N$ and number of features $D$ are listed in Table \ref{table:uci}. For each of the datasets, we use the 20 random train-test splits provided by \citet{yarin16dropout}\footnote{The splits are publicly available from \url{https://github.com/yaringal/DropoutUncertaintyExps}}. 
Following earlier work, we use 30 iterations of Bayesian Optimization (BO) to tune the prior precision $\lambda$ and the noise precision $\tau$. For each iteration of BO, 5-fold cross-validation is used to evaluate the considered hyperparameter setting. This is repeated for each of the 20 train-test splits for each dataset.
The final values reported in the table for each dataset are the mean and standard error from these 20 runs. The final runs for the 8 datasets are shown in Figure~\ref{figure:uci_regression_full}. 

Following earlier work, we use neural networks with one hidden layer and 50 hidden units with ReLU activation functions. All networks were trained for 40 epochs.
For the 4 smallest datasets, we use a minibatch size of 32, 10 MC samples for Vadam and 20 MC samples for BBVI. 
For the 4 larger datasets, we use a minibatch size of 128, 5 MC samples for Vadam and 10 MC samples for BBVI.
For evaluation, 100 MC samples were used in all cases.

For BBVI, we optimize the variational objective using the Adam optimizer.
For both BBVI and Vadam we use a learning rate of $\alpha = 0.01$ and set $\gamma_1 = 0.99$ and $\gamma_2 = 0.9$ to encourage convergence within 40 epochs.
For both BBVI and Vadam, the initial precision of the variational distribution $q$ was set to 10.

\subsection{VOGN Convergence Experiments}

We apply BBVI, Vadam, and VOGN to train a neural network with a single-hidden layer of 64 units and ReLU activations on a random train-test split of the Australian-Scale dataset ($N=690$, $D=14$). For VOGN, we do not use the natural-momentum term. The prior precision $\lambda$ is set to $1$. We run each method for 5000 iterations. For both Adam and Vadam, we set $\alpha = 0.001$, $\gamma_1 = 0.9$, and $\gamma_2=0.999$. For VOGN, we set $\alpha=0.001$ and $\gamma_1=0.9$. We run experiments for different minibatch sizes $M$ and number of MC samples $S$. The left side of Figure \ref{figure:results} shows results for $(M=1, S=1)$, $(M=1, S=16)$ and $(M=128, S=16)$.

\section{Details for the Exploration for Deep Reinforcement Learning Experiment}
\label{app:rl}
Reinforcement learning (RL) aims to solve the sequential decision making problem where at each discrete time step $t$ an agent observes a state $\state_t$ and selects an action $\action_t$ using a policy $\pi$, i.e., $\action_t \sim \pi(\action|\state_t)$. 
The agent then receives an immediate reward $r_t = r(\state_t,\action_t)$ and observes a next state $\state_t \sim p(\state'|\state_t, \action_t)$.
The goal in RL is to learn the optimal policy $\pi^*$ which maximizes the expected return $\mathbb{E}\left[ \sum_{t}^\infty \gamma^{t-1} r_t \right]$ where $\gamma$ is the discounted factor and the expectation is taken over a sequence of densities $\pi(\action|\state_t)$ and $p(\state'|\state_t, \action_t)$.

A central component of RL algorithms is the Q-function, $Q^{\pi}(\state,\action)$, which denotes the expected return after executing an action $\action$ in a state $\state$ and following the policy $\pi$ afterwards. 
Formally, the Q-function is defined as 
$
Q^{\pi}(\state,\action) = \mathbb{E}\left[ \sum_{t=1}^\infty \gamma^{t-1} r_t | \state_1 = \state, \action_1 = \action\right]
$.
The Q-function also satisfies a recursive relation also known as the Bellman equation:
$
Q^{\pi}(\state,\action) = r(\state,\action) + \gamma \mathbb{E}_{p(\state'|\state,\action) \pi(\action'|\state')} \left[ Q^{\pi}(\state', \action') \right]
$.
Using the Q-function and a parameterized policy $\pi_{\theta}$, the goal of reinforcement learning can be simply stated as finding a policy parameter $\vtheta$ which maximizes the expected Q-function\footnote{In this section, we omit a case where the state distribution $p(\state)$ depends on the policy.
	In practice many policy gradient methods (especially actor-critic type methods) also often ignore the dependency between the state distribution and the policy.}:
\begin{align}
\max_{\params} \mathbb{E}_{p(\state) \pi_{\params}(\action|\state)}\left[  Q^{\pi}(\state,\action) \right]. \label{eq:rl_objective_ori}
\end{align}

In practice, the Q-function is unknown and is commonly approximated by a parameterized function $\widehat{Q}_{\omega}(\state, \action)$ with parameter $\vomega$ learned such that it satisfies the Bellman equation on average:
$
\min_{\vomega} \mathbb{E}_{p(\state) \beta(\action|\state) }[ ( r(\state,\action) + \gamma \mathbb{E}_{\pi_{\params}(\action'|\state')} [ \widehat{Q}_{\tilde{\vomega}}(\state', \action') ] - \widehat{Q}_{\vomega}(\state,\action) )^2  ]
$,
where $\beta(\va|\vs)$ is a behavior policy used to collect samples and $\tilde{\vomega}$ is either a copy or a slowly updated value of $\vomega$ whose $\nabla_{\vomega} \widehat{Q}_{\tilde{\vomega}}(\state', \action') = 0$.
By using an approximated Q-function, the goal of RL is to find a policy parameter maximizing the expected value of $\widehat{Q}_{\vomega}(\state, \action)$:
\begin{align}
\max_{\params} \mathbb{E}_{p(\state) \pi_{\params}(\action|\state)}\left[  \widehat{Q}_{\vomega}(\state,\action) \right]
:= \min_{\params} -\mathbb{E}_{p(\state) \pi_{\params}(\action|\state)}\left[  \widehat{Q}_{\vomega}(\state,\action) \right] := \min_{\params} F(\params)
. \label{eq:rl_q_objective}
\end{align}
In the remainder, we consider the minimization problem $\min_{\params} F(\params)$ to be consistent with the variational optimization problem setting in the main text.

\subsection{Stochastic Policy Gradient and Deterministic Policy Gradient}

The RL objective in~\eqref{eq:rl_q_objective} is often minimized by gradient descent.
The gradient computation depends on stochasticity of $\pi_{\params}$.
For a stochastic policy, $\pi_{\params}(\action|\state)$, \emph{policy gradient} or REINFORCE can be computed using the likelihood ratio trick:
\begin{align}
F(\params) = -\mathbb{E}_{p(\state) \pi_{\params}(\action|\state)}\left[  \widehat{Q}_{\vomega}(\state,\action) \right], \quad
\nabla_{\params} F(\params) = -\mathbb{E}_{p(\state) \pi_{\params}(\action|\state)} \left[ \nabla_{\params} \log \pi_{\params}(\action|\state) \widehat{Q}_{\vomega}(\state,\action) \right]. \label{eq:spg}
\end{align}
For a deterministic policy $\pi_{\params}(\state)$, \emph{deterministic policy gradient} (DPG)~\citep{DBLP:conf/icml/SilverLHDWR14} can be computed using the chain-rule:
\begin{align}
F(\params) = -\mathbb{E}_{p(\state)}\left[  \widehat{Q}_{\vomega}(\state, \pi_{\params}(\state)) \right], \quad
\nabla_{\params} F(\params) = -\mathbb{E}_{p(\state)} \left[ \nabla_{\params} \pi_{\params}(\state) \nabla_{\action} \widehat{Q}_{\vomega}(\state,\pi_{\params}(\state)) \right]. \label{eq:dpg}
\end{align}
As discussed by~\citet{DBLP:conf/icml/SilverLHDWR14}, the deterministic policy gradient is more advantageous than the stochastic counter part due to its lower variance.
However, the issue of a deterministic policy is that it does not perform \emph{exploration} by itself.
In practice, exploration is done by injecting a noise to the policy output, i.e., $\action = \pi_{\params}(\state) + \boldsymbol{\epsilon}$ where $\boldsymbol{\epsilon}$ is a noise from some random process such as Gaussian noise.
However, action-space noise may be insufficient in some problems~\citep{DBLP:journals/paladyn/RuckstiessSSWSS10}. 
Next, we discussed parameter-based exploration approach where exploration is done in the \emph{parameter space}.
Then, we show that such exploration can be achieved by simply applying VadaGrad and Vadam to policy gradient methods.

\subsection{Parameter-based Exploration Policy Gradient}

\emph{Parameter-based exploration policy gradient}~\citep{DBLP:journals/paladyn/RuckstiessSSWSS10} relaxes the RL objective in~\eqref{eq:rl_q_objective} by assuming that the parameter $\params$ is sampled from a Gaussian distribution $q(\params) := \mathcal{N}(\params | \vmu, \vsigma^2)$ with a diagonal covariance.
Formally, it solves an optimization problem 
\begin{align}
\min_{\boldsymbol{\mathrm{\mu}},\boldsymbol{\mathrm{\sigma^2}}} \,\, \myexpect_{\mathcal{N}(\params|\boldsymbol{\mathrm{\mu}},\boldsymbol{\mathrm{\sigma}}^2)}[F(\params)], \label{eq:pgpe}
\end{align}
where $F(\params)$ is either the objective function for the stochastic policy in~\eqref{eq:spg} or the deterministic policy in~\eqref{eq:dpg}.
In each time step, the agent samples a policy parameter $\params \sim \mathcal{N}(\params| \vmu, \vsigma^2)$ and uses it to determine an action\footnote{The original work of \cite{DBLP:journals/paladyn/RuckstiessSSWSS10} considers an episode-based method where the policy parameter is sampled only at the start of an episode. However, we consider DPG which is a step-based method. Therefore, we sample the policy parameter in every time step. Note that we may only sample the policy parameter at the start of the episode as well.}.
This exploration strategy is advantageous since the stochasticity of $\params$ allows the agent to exhibit much more richer explorative behaviors when compared with exploration by action noise injection.

Notice that~\eqref{eq:pgpe} is exactly the variational optimization problem discussed in the main text.
As explained in the main text, this problem can be solved by our methods.
In the next section, we apply VadaGrad and Vadam to the deep deterministic policy gradient and show that a parameter-exploration strategy induced by our methods allows the agent to achieve a better performance when compared to existing methods.

\subsection{Parameter-based Exploration Deep Deterministic Policy Gradient via VadaGrad and Vadam}

While parameter-based exploration strategy can be applied to both stochastic and deterministic policies, it is commonly applied to a deterministic policy.
In our experiment, we adopt a variant of deterministic policy gradient called \emph{deep deterministic policy gradient}~(DDPG)~\citep{DBLP:journals/corr/LillicrapHPHETS15}.
In DDPG, the policy $\pi_{\params}(\state)$ and the Q-function $\widehat{Q}_{\vomega}(\state, \action)$ are represented by deep neural networks.
To improve stability, DDPG introduces target networks, $\pi_{\tilde{\params}}(\state)$ and $\widehat{Q}_{\tilde{\vomega}}(\state, \action)$, whose weight parameters are updated by $\tilde{\params} \leftarrow (1 - \tau)\tilde{\params} + \tau \params $ and $\tilde{\vomega} \leftarrow (1 - \tau)\tilde{\vomega} + \tau \vomega $ for $0 < \tau < 1$.
The target network are used to update the Q-function by solving
\begin{align}
\min_{\vomega} \mathbb{E}_{p(\state) \beta(\action|\state) }\left[ \left( r(\state,\action) + \gamma \widehat{Q}_{\tilde{\vomega}}(\state', \pi_{\tilde{\params}}(\state')) - \widehat{Q}_{\vomega}(\state,\action) \right)^2  \right].
\label{eq:q_residual_deep}
\end{align}
Gradient descent on~\eqref{eq:q_residual_deep} yields an update: $\vomega \leftarrow \vomega + \kappa \mathbb{E}_{p(\state) \beta(\action|\state) }\left[ \left( r(\state,\action) + \gamma \widehat{Q}_{\tilde{\vomega}}(\state', \pi_{\tilde{\params}}(\state')) - \widehat{Q}_{\vomega}(\state,\action) \right) \nabla_{\vomega} \widehat{Q}_{\vomega}(\state,\action) \right]$, where $\kappa > 0$ is the step-size.
DDPG also uses a \emph{replay buffer} which is a first-in-first-out queue that store past collected samples.
In each update iteration, DDPG uniformly draws $M$ minibatch training samples from the replay buffer to approximate the expectations.

To apply VadaGrad to solve~\eqref{eq:pgpe}, in each update iteration we sample $\params_t \sim \mathcal{N}(\params|\vmu_t, \vsigma^2_t)$ and then updates the mean and variance using the VadaGrad update in~\eqref{eq:VadaGrad}.
The deterministic policy gradient $\nabla_{\params} F(\params)$ can be computed using the chain-rule as shown in~\eqref{eq:dpg}.
The computational complexity of DDPG with VadaGrad is almost identical to DDPG with Adagrad, except that we require gradient computation at sampled weight $\params_t$.
Algorithm \ref{algorithm:dpg_vadagrad} below outlines our \emph{parameter-based exploration DPG via VadaGrad} where the only difference from DDPG is the sampling procedure in line 3.
Note that we use $N = 1$ in this case.
Also note that the target policy network $\pi_{\tilde{\boldsymbol{\mathrm{\mu}}}}$ does not performed parameter-exploration and it is updated by the mean $\vmu$ instead of a sampled weight $\params$.

In VadaGrad, the precision matrix always increases overtime and it guarantees that the policy eventually becomes deterministic.
This is beneficial since it is known that there always exists a deterministic optimal policy for MDP.
However, this behavior may not be desirable in practice since the policy may become deterministic too fast which leads to premature convergence.
Moreover, for VadaGrad the effective gradient step-size, $\frac{\alpha}{\sqrt{\vs_{t+1}}}$, will be close to zero for a nearly deterministic variational distribution $q$ which leads to no policy improvement.
This issue can be avoided by applying Vadam instead of VadaGrad.
As will be shown in the experiment, Vadam allows the agent to keep explore and avoids the premature convergence issue.
Note that parameter-based exploration RL is not a VI problem and we require some modification to the Vadam update, as shown in line 3 of Algorithm~\ref{algorithm:dpg_vadam} below.

We perform experiment using the Half-Cheetah task from the OpenAI gym platform.
We compare DDPG with \textbf{VadaGrad} and \textbf{Vadam} against four baseline methods.
\begin{itemize}
\item \textbf{SGD-Plain:} the original DDPG without any noise injection optimized by SGD,
\item \textbf{Adam-Plain:} the original DDPG without any noise injection optimized by Adam,
\item \textbf{SGD-Explore:} a naive parameter exploration DDPG based on VO optimized by SGD, and
\item \textbf{Adam-Explore:} a naive parameter exploration DDPG based on VO optimized by Adam.
\end{itemize}
In SGD-Explore and Adam-Explore, we separately optimizes the mean and variance of the Gaussian distribution:
\begin{align}
\vmu_{t+1} &= \vmu_t - \alpha \nabla_{\boldsymbol{\mu}} \myexpect_{\mathcal{N}(\params|\boldsymbol{\mathrm{\mu}_t},\boldsymbol{\mathrm{\sigma}}^2_t)}[F(\params)], \quad\quad
\vsigma^2_{t+1} = \vsigma^2_t - \alpha^{(\sigma)} \nabla_{\boldsymbol{\sigma}^2} \myexpect_{\mathcal{N}(\params|\boldsymbol{\mathrm{\mu}_t},\boldsymbol{\mathrm{\sigma}}^2_t)}[F(\params)],
\end{align}
where $\alpha > 0$ is the mean step-size and $\alpha^{(\sigma)} > 0$ is the variance step-size. 
The gradients of $\myexpect_{\mathcal{N}(\params|\boldsymbol{\mathrm{\mu}_t},\boldsymbol{\mathrm{\sigma}}^2_t)}[F(\params)]$ are computed by chain-rule and automatic-differentiation.
For Adam-Explore, two Adam optimizers with different scaling vectors are used to independently update the mean and variance.

All methods use the DDPG network architectures as described by~\citet{DBLP:journals/corr/LillicrapHPHETS15};
two-layer neural networks with $400$ and $300$ ReLU hidden units.
The output of the policy network is scaled by a hyperbolic tangent to bound the actions.
The minibatch size is $M=64$.
All methods optimize the Q-network by Adam with step-size $\kappa = 10^{-3}$.
The target Q-network and target policy network use a moving average step-size $\tau = 10^{-3}$.
The expectation over the variational distribution of all methods are approximated using one MC sample.
For optimizing the policy network, we use the step-sizes given in Table~\ref{table:rl_setting}.
\begin{table}[t]
\centering
\begin{tabular}{ l | l || c | c | c | c | c | c } 
\multicolumn{2}{c || }{Method} 			& $\alpha$		& $\gamma_2$	& $\gamma_1$	& $\alpha^{(\sigma)}$	& $\gamma_2^{(\sigma)}$	& $\gamma_1^{(\sigma)}$ \\ \hline\hline
\multirow{2}{*}{VO/VI} 	& {VadaGrad}	& $10^{-2}$		& $0.99$	& -			& -				& -				& - \\ 	\cline{2-8}
						& {Vadam}		& $10^{-4}$		& $0.999$	& $0.9$		& -				& -				& - \\ \hline\hline
\multirow{2}{*}{Plain}	& SGD-Plain 	& $10^{-4}$		& -			& -			& -				& -				& - \\ \cline{2-8}
						& Adam-Plain	& $10^{-4}$		& $0.999$	& $0.9$		& -				& -				& - \\ \hline\hline
\multirow{2}{*}{Explore}& SGD-Explore 	& $10^{-4}$		& -			& -			& $10^{-2}$		& -				& - \\ \cline{2-8}
						& Adam-Explore	& $10^{-4}$		& $0.999$	& $0.9$		& $10^{-2}$		& $0.999$		& $0.9$ \\ \hline	
\end{tabular}
\caption{Hyper-parameter setting for the deep reinforcement learning experiment. We refer to Algorithm~\ref{algorithm:dpg_vadagrad} and~\ref{algorithm:dpg_vadam} for the meaning of each hyper-parameter.}
\label{table:rl_setting}
\end{table}
For Adam we use $\delta = 10^{-8}$. We also use the same value of $\lambda = 10^{-8}$ for Vadam.
The initial precision for SGD-Explore, Adam-Explore, and VadaGrad is $\vsigma_{t=1}^{-2} = 10000$. 

For Vadam we use $\vs_{t=1} = 0$ for the initial second-order moment estimate and add a constant value of $c=10000$ to the precision matrix $\vsigma_{t}^{-2} = \vs_{t} + \lambda + c$ for sampling (see line 3 of Algorithm~\ref{algorithm:dpg_vadam}).
We do this for two reasons.
First, we set $\vs_{t=1} = 0$ so that the initial conditions and hyper-parameters of Vadam and Adam-Plain are exactly the same.
Second, we add a constant $c$ to prevent an ill-conditioned precision matrix during sampling.
Without this constant, the initial precision value of $\vsigma_{t=1}^{-2} = \lambda = 10^{-8}$ is highly ill-conditioned and sampled weights do not contain any information.
This is highly problematic in RL since we do not have training samples at first and the agent needs to collect training samples from scratch. 
Training samples collected initially using $\vsigma_{t=1}^{-2} = 10^{-8}$ is highly uninformative (e.g., all actions are either the maximum or minimum action values) and the agent cannot correctly estimate uncertainty.
We emphasize that the constant is only used for sampling and not used for gradient update.
We expect that this numerical trick is not required for a large enough value of $\lambda$, but setting an appropriate value of $\lambda$ is not trivial in deep learning and is not in the scope of this paper.
As such, we leave finding a more appropriate approach to deal with this issue as a future work.

\begin{figure}[p]
	\centering
	\includegraphics[width=0.70\linewidth]{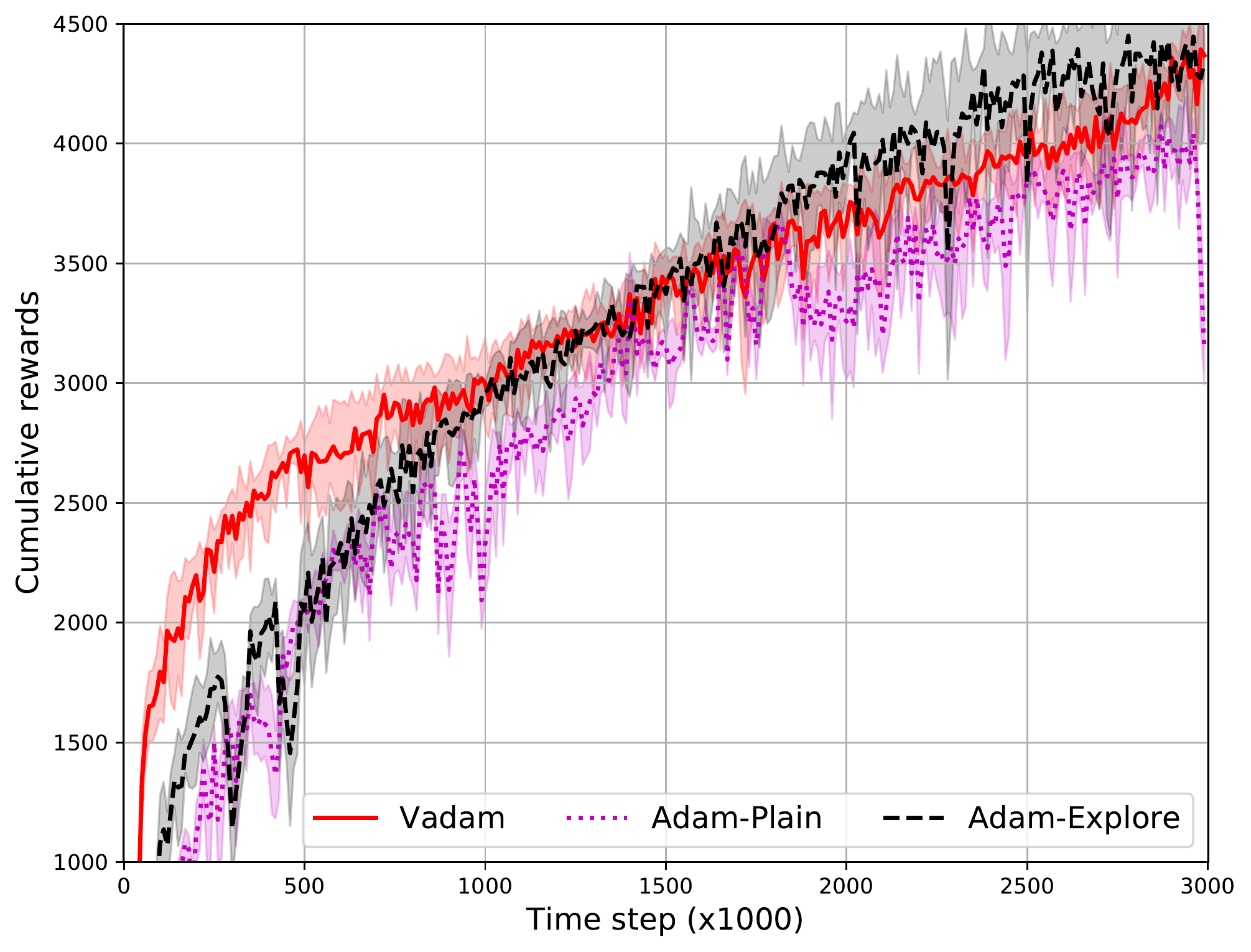}
	\caption{The early learning performance of Vadam, Adam-Plain and Adam-Explore, on the half-cheetah task in the reinforcement learning experiment. Vadam shows faster learning in this early stage of learning. The mean and standard error are computed over 5 trials.}
	\label{figure:rl_exp_zoom}
\end{figure}

We perform experiment using the Half-Cheetah task from the OpenAI gym platform~\citep{1606.01540}.
We measure the performance of each method by computing cumulative rewards along 20 test episodes without exploration. 
The early learning performance of Vadam, Adam-Plain, and Adam-Explore in Figure~\ref{figure:rl_exp_zoom} shows that Vadam learns faster than the other methods.
We conjecture that exploration through a variational distribution allows Vadam agent to collect more information training samples when compare to Adam-Plain.
While Adam-Explore also performs exploration through a variational distribution, its performance is quite unstable with high fluctuations.
This fluctuation in Vadam-Explore is likely because the mean and variance of the variational distribution are optimized independently.
In contrast, Vadam uses natural-gradient to optimizes the two quantities in a strongly correlated manner, yielding a more stable performance.

\begin{figure}[p]
	\centering
	\includegraphics[width=0.70\linewidth]{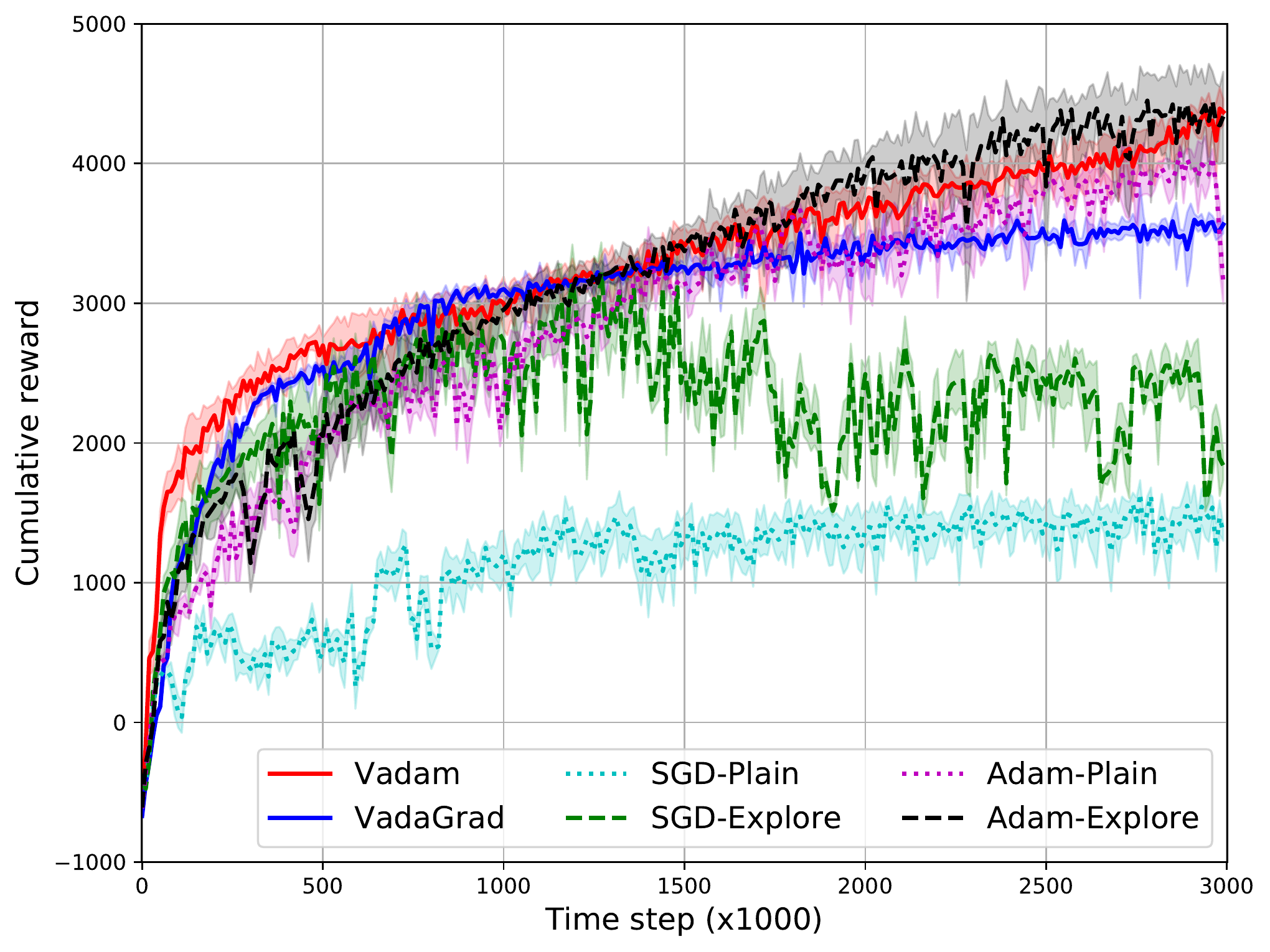}
	\caption{The performance of all evaluated methods on the half-cheetah task in the reinforcement learning experiment. 
		Vadam and Adam-based methods perform well overall and give comparable final performance. 
		VadaGrad also learns well but shows sign of premature convergence.
		SGD-based method do not learn well throughout.
		The mean and standard error are computed over 5 trials.}
	\label{figure:rl_exp}
\end{figure}

Figure~\ref{figure:rl_exp} shows the learning performance for a longer period of training for all methods.
We can see that VadaGrad learns faster than SGD and Adam-based methods initially, but it suffers from a premature convergence and are outperformed by Adam-based methods.
In contrast, Vadam does not suffer from the premature convergence.
We can also observe that while Adam-based method learn slower than Vadam initially, they eventually catch up with Vadam and obtain a comparable performance at the 3 million time-steps.
We conjecture that this is because exploration strategy is very important in the early stage of learning where the agent does not have sufficient amount of informative training samples. 
As learning progress and there is a sufficient amount of informative samples, exploration would not help much.
Nonetheless, we can still see that Vadam and Adam-Explore give slightly better performance than Adam-Plain, showing that parameter-based exploration is still beneficial for DDPG.

\begin{algorithm}[p]
	\begin{algorithmic}[1]
		\STATE \textbf{Initialize:} Variational distribution $\mathcal{N}(\params | \vmu_{1}, \vs^{-1}_{1})$ with random initial mean and initial precision $\vs_1 = 10000$.
		
		\FOR {Time step $t = 1, ..., \infty$}
		\STATE Sample policy parameter $\vtheta_t \sim \mathcal{N}(\vtheta | \vmu_t, \vs^{-1}_t)$.
		\STATE Observe $\state_t$, execute $\action_t = \pi_{\params_t}(\state_t)$ observe $r_t$ and transit to $\state'_t$. Then add $(\state_t, \action_t, r_t, \state'_t)$ to a replay buffer $\mathcal{D}$.
		\STATE Drawn M minibatch samples $\{(\state_i, \action_i, r_i, \state'_i)\}_{i=1}^M$ from $\mathcal{D}$.
		\STATE Update the Q-network weight $\vomega$ by stochastic gradient descent or Adam:
		\begin{align*}
		\vomega_{t+1} = \vomega_t + {\kappa}\textstyle \sum_{i=1}^M \left( r_i + \gamma \widehat{Q}_{\tilde{\vomega}_t}(\state_i', \pi_{\tilde{\boldsymbol{\mathrm{\mu}}}_t}(\state_i')) - \widehat{Q}_{\vomega_t}(\state_i,\action_i) \right) \nabla_{\vomega} \widehat{Q}_{\vomega_t}(\state_i,\action_i) / M .
		\end{align*}
		\STATE Compute deterministic policy gradient using the sampled policy parameter:
		\begin{align*}
		\hat{\nabla}_{\params} F(\params_t) = -\textstyle \sum_{i=1}^M \nabla_{{\boldsymbol{\mathrm{\theta}}}}\pi_{{\boldsymbol{\mathrm{\theta}}}_t}(\state_i) \nabla_{\action} Q_{\vomega_{t+1}}(\state_i, \pi_{{\boldsymbol{\mathrm{\theta}}}_t}(\state_i) ) / M.
		\end{align*}
		
		\STATE Update the mean $\vmu$ and variance $\vsigma^2$ by VadaGrad:
		\begin{align*}
		\vmu_{t+1} = \vmu_{t} - \alpha \hat{\nabla}_{\params} F(\params_t) /\sqrt{\vs_{t+1}}, \quad
		\vs_{t+1} = \vs_{t} + (1-\gamma_2) [ \hat{\nabla}_{\params} F(\params_t) ]^2.
		\end{align*}
		
		\STATE Update target network parameters $\tilde{\vomega}_{t+1}$ and $\tilde{\vmu}_{t+1}$ by moving average:
		\begin{align*}
		\tilde{\vomega}_{t+1} = (1 - \tau) \tilde{\vomega}_{t} + \tau \vomega_{t+1}, \quad
		\tilde{\vmu}_{t+1} = (1 - \tau) \tilde{\vmu}_t + \tau \vmu_{t+1}.
		\end{align*}
		\ENDFOR
	\end{algorithmic} 
	\caption{Parameter-based exploration DDPG via VadaGrad}
	\label{algorithm:dpg_vadagrad}
\end{algorithm}
\begin{algorithm}[p]
	\begin{algorithmic}[1]
		\STATE \textbf{Initialize:} Initial mean $\vmu_1$, 1st-order moment $\vm_1 = 0$, 2nd-order moment $\vs_1=0$, prior $\lambda=10^{-8}$, constant $c=10000$.
		
		\FOR {Time step $t = 1, ..., \infty$}
		\STATE Sample policy parameter $\vtheta_t \sim \mathcal{N}(\vtheta | \vmu_t, (\vs_t + \lambda + c )^{-1} )$.
		\STATE Observe $\state_t$, execute $\action_t = \pi_{\params_t}(\state_t)$ observe $r_t$ and transit to $\state'_t$. Then add $(\state_t, \action_t, r_t, \state'_t)$ to a replay buffer $\mathcal{D}$.
		\STATE Drawn M minibatch samples $\{(\state_i, \action_i, r_i, \state'_i)\}_{i=1}^M$ from $\mathcal{D}$.
		\STATE Update the Q-network weight $\vomega$ by stochastic gradient descent or Adam:
		\begin{align*}
		\vomega_{t+1} = \vomega_t + {\kappa} \textstyle \sum_{i=1}^M \left( r_i + \gamma \widehat{Q}_{\tilde{\vomega}_t}(\state_i', \pi_{\tilde{\boldsymbol{\mathrm{\mu}}}_t}(\state_i')) - \widehat{Q}_{\vomega_t}(\state_i,\action_i) \right) \nabla_{\vomega} \widehat{Q}_{\vomega_t}(\state_i,\action_i) / M.
		\end{align*}
		\STATE Compute deterministic policy gradient using the sampled policy parameter:
		\begin{align*}
		\hat{\nabla}_{\params} F(\params_t) = -\textstyle \sum_{i=1}^M \nabla_{{\boldsymbol{\mathrm{\theta}}}}\pi_{{\boldsymbol{\mathrm{\theta}}}_t}(\state_i) \nabla_{\action} Q_{\vomega_{t+1}}(\state_i, \pi_{{\boldsymbol{\mathrm{\theta}}}_t}(\state_i) ) / M.
		\end{align*}
		
		\STATE Update and correct the bias of the 1st-order moment $\vm$ and the 2nd-order moment $\vs$ by Vadam:
		\begin{align*}
		\vm_{t+1} &= \gamma_1 \vm_{t} + (1-\gamma_1) (\hat{\nabla}_{\params} F(\params_t) + \lambda\vmu_t), \quad
		&&\vs_{t+1} = \gamma_2 \vs_{t} + (1-\gamma_2) [ \hat{\nabla}_{\params} F(\params_t) ]^2, \\
		\hat{\vm}_{t+1} &= \vm_t / ( 1 - \gamma_1^t), \quad 
		&&\hat{\vs}_{t+1} = \vs_{t+1} / ( 1 - \gamma_2^t).
		\end{align*}
		
		\STATE Update the mean $\vmu$ using the moment estimates by Vadam: 
		\begin{align*}
		\vmu_{t+1} = \vmu_t - \alpha \hat{\vm}_{t+1} / (\sqrt{\hat{\vs}_t} + \lambda).
		\end{align*}

		\STATE Update target network parameters $\tilde{\vomega}_{t+1}$ and $\tilde{\vmu}_{t+1}$ by moving average:
		\begin{align*}
		\tilde{\vomega}_{t+1} = (1 - \tau) \tilde{\vomega}_{t} + \tau \vomega_{t+1}, \quad
		\tilde{\vmu}_{t+1} = (1 - \tau) \tilde{\vmu}_t + \tau \vmu_{t+1}.
		\end{align*}
		\ENDFOR
	\end{algorithmic} 
	\caption{Parameter-based exploration DDPG via Vadam}
	\label{algorithm:dpg_vadam}
\end{algorithm}


\section{Toy Example on Local-Minima Avoidance using Vadam}
\label{app:3d_illustration}

\begin{figure}[!t]
\centering
\includegraphics[scale=0.5]{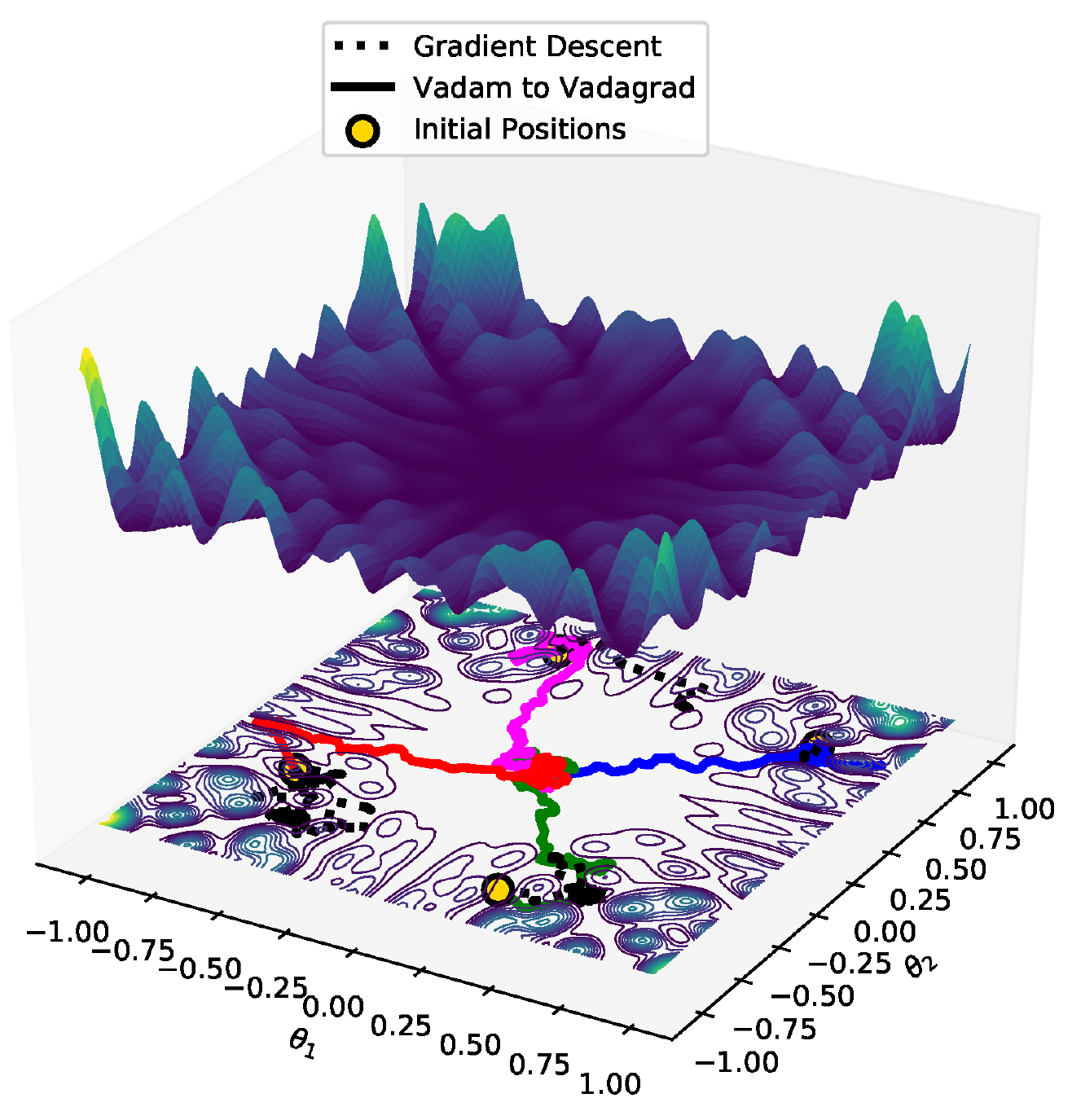}
\caption{Illustration of variational optimization on a complex 2D objective function. Variational optimization is performed from four different initial positions. The four runs are shown in solid lines in different colors. Gradient descent (shown in dotted, black lines) is also initialized at the same locations. `Vadam to VadaGrad' shows ability to navigate the landscape to reach the flat (and global) minimum, while gradient descent gets stuck in various locations.}
\label{fig:3d_illustration}
\end{figure}

Fig. \ref{fig:3d_illustration} shows an illustration of variational optimization on a two-dimensional objective function.
The objective function $h(x,y) = \exp \{-(x\sin(20y)+y\sin(20x))^2 - (x\cos(10y)-y\sin(10x))^2\}$ is taken from Fig. 5.2 in \citet{robert2005}. Variational optimization is performed by gradually turning off the KL-term for Vadam, thus annealing Vadam towards VadaGrad. This is referred to as ``Vadam to VadaGrad".
We show results for 4 multiple runs of each method started with a different initial values.
The figure shows that variational optimization can better navigate the landscape to reach the flat (and global) minimum than gradient descent.

\section{Experiment on Improving ``Marginal Value of Adaptive-Gradient Methods"}
\label{sec:rnn}
\begin{figure}[t]
\centering
\includegraphics[width=4.0in]{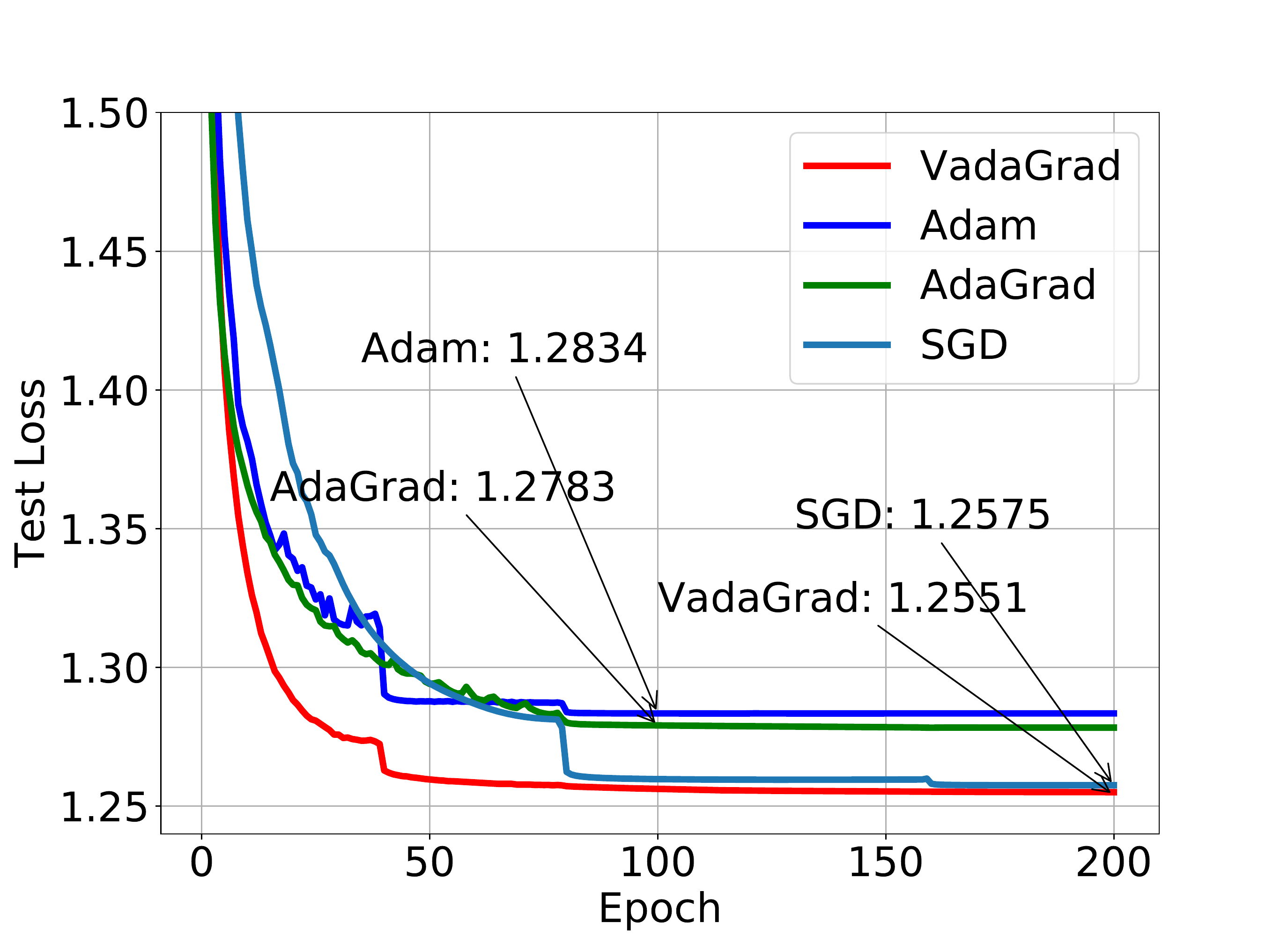} 
\caption{Results for character-level language modeling using the War and Peace dataset for the ``Marginal Value of Adaptive-Gradient Methods'' experiment. We repeat the results shown in Fig. 2 (b) of \citet{wilson2017marginal} and show that VadaGrad does not suffer from the issues pointed in that paper, and it performs comparable to SGD.}
\label{fig:rnn_results}
\end{figure}

Recently, \citet{wilson2017marginal} show some examples where adaptive gradient methods, namely Adam and AdaGrad, generalize worse than SGD. 
We repeated their experiments to see whether weight-perturbation in VadaGrad improves the generalization performance of AdaGrad. 
We firstly consider an experiment on the character-level language modeling on the novel War and Peace dataset~(shown in Fig. 2b in \citet{wilson2017marginal}).
Figure~\ref{fig:rnn_results} shows the test error of SGD, AdaGrad, Adam and VadaGrad.
We can see that VadaGrad generalizes well and achieves the same performance as SGD unlike AdaGrad and Adam. 
We also repeated the CIFAR-10 experiment discussed in the paper, and found that the improvement using VadaGrad was minor. We believe that regularization techniques such as batch normalization, batch flip, and dropout, play an important role for the CIFAR-10 dataset and that is why we did not see an improvement by using VadaGrad.
Further investigation will be done in future works.

We use the following hyper-parameter setting in this experiment.
For all methods, we divide the step-size $\alpha$ by 10 once every $K$ epochs, as described by~\citet{wilson2017marginal}.
For VadaGrad, AdaGrad, and Adam, we fixed the value of scaling vector step-size $\beta$ and do not decay.
For AdaGrad and Adam, the initial value of scaling vector is 0. 
For VadaGrad, we use the initial value $\vs_1 = 10^{-4}$ to avoid numerical issue of sampling from Gaussian with 0 precision. (Recall that VadaGrad does not have a prior $\lambda$).
We perform grid search to find the best value of initial $\alpha$, $K$, and $\beta$ for all methods except Adam which use the default value of $\beta = 0.999$ and $\gamma = 0.9$. 
The best hyper-parameter values which give the minimum test loss are given in Table~\ref{table:marginal}.

\begin{table}[H]
\centering
\begin{tabular}{ l | c | c | c } 
\textbf{Method}	& $K$ 	& $\alpha$ 	& $\beta$	\\ \hline \hline
VadaGrad 	& 40 	& 0.0075 	& 0.5			\\ \hline
SGD			& 80	& 0.5		& -				\\ \hline
AdaGrad		& 80	& 0.025		& 0.5			\\ \hline
Adam		& 40	& 0.0012	& 0.999			\\
\end{tabular}
\caption{Hyper-parameter setting for the ``Marginal Value of Adaptive-Gradient Methods'' experiment.}
\label{table:marginal}
\end{table}

\end{document}